\documentclass[5p,times]{elsarticle}

\usepackage[utf8]{inputenc}
\usepackage[T1]{fontenc}

\usepackage{amsmath,amssymb,amsfonts}
\usepackage{algorithm}
\usepackage{algorithmic}
\usepackage{graphicx}
\graphicspath{{figures/figures/}{figures/}{./}}
\usepackage{textcomp}
\usepackage{xcolor}
\usepackage{booktabs}
\usepackage{multirow}

\usepackage{bm}
\usepackage{xurl}
\usepackage[protrusion=false]{microtype}
\usepackage{ragged2e}
\usepackage{hyperref}
\usepackage{doi}
\usepackage{makecell}
\usepackage{tabularx}
\usepackage{tikz}
\usetikzlibrary{positioning,arrows.meta,fit,calc}

\hypersetup{
    colorlinks=true,
    allcolors=blue,
    breaklinks=true,
    linkcolor=blue,
    citecolor=blue,
    urlcolor=blue
}

\def\BibTeX{{\rm B\kern-.05em{\sc i\kern-.025em b}\kern-.08em T\kern-.1667em\lower.7ex\hbox{E}\kern-.125emX}}

\newcommand{\NR}{\textnormal{NR}}

\usepackage{multicol}
\usepackage{etoolbox}
\makeatletter
\patchcmd{\highlights}{\begin{itemize}}{\setlength{\linewidth}{\textwidth}\begin{multicols}{2}\raggedcolumns\interlinepenalty=10000 \begin{itemize}}{}{}
\patchcmd{\endhighlights}{\end{itemize}}{\end{itemize}\end{multicols}}{}{}
\makeatother
\journal{Knowledge-Based Systems}
\begin{document}
\raggedbottom
\setlength{\emergencystretch}{4em}

\begin{frontmatter}
\title{Leakage-Controlled Multimodal Learning for Diagnosis and Progression Prediction in Alzheimer's Disease Research}
\author[lissi]{Akeem Temitope Otapo}
\ead{akeem.otapo@u-pec.fr}
\author[lissi]{Ghazaleh Khodabandelou\corref{cor1}}
\ead{ghazaleh.khodabandelou@u-pec.fr}
\cortext[cor1]{Corresponding author.}

\author[l2ti]{Zuheng Ming}
\ead{zuheng.ming@univ-paris13.fr}
\author[lissi]{Alice Othmani}
\ead{alice.othmani@u-pec.fr}

\affiliation[lissi]{organization={Laboratoire Images, Signaux et Systèmes Intelligents (LiSSi), Université Paris-Est Créteil (UPEC)},
            addressline={122 rue Paul Armangot},
            city={Vitry-sur-Seine},
            postcode={94400},
            country={France}}
\affiliation[l2ti]{organization={L2TI, Institut Galilée, Université Sorbonne Paris Nord (USPN)},
            addressline={99 avenue Jean-Baptiste Clément},
            city={Villetaneuse},
            postcode={93430},
            country={France}}
\begin{abstract}
Alzheimer's disease prediction involves irregular visits, heterogeneous measurements and incomplete modalities. This study presents a multimodal multitask framework combining an adapted SFCN MRI encoder, four causal clinical Transformers, shared fusion and task-specific ODE-GRU dynamics. Fine-tuning and LoRA adapt the final two MRI blocks. Task-DRO balances task losses, while Group-CVaR targets cohort and comorbidity strata. Branch-specific input controls, subject-grouped partitions and empirical causality checks support longitudinal evaluation.
Across 2,649 subjects and 17,317 visits from ADNI, OASIS-2 and MIRIAD, internal validation yields diagnosis, stage-1 progression and first-stage-1-visit progression AUROCs of 0.935$\pm$0.002, 0.884$\pm$0.003 and 0.870$\pm$0.005, respectively (mean$\pm$SD across three seeds). Corresponding hybrid AUROCs are 0.951, 0.909 and 0.896. Next-visit MMSE mean absolute error (MAE) is 1.61 points; worst-stratum diagnosis AUROC is 0.827$\pm$0.008. Sampled ADNI explanations identify task-specific input dependence. OASIS-3 external validation yields network and hybrid diagnosis AUROCs of 0.763 and 0.767, hybrid next-visit progression AUROC of 0.764, diagnosis calibration error decreasing from 0.197 to 0.052, and next-visit MMSE MAE of 0.86. Seed-42 paired ablations of six components yield pooled diagnosis and progression AUROC differences between $-0.004$ and $+0.004$; removing clinical encoder inputs lowers diagnosis AUROC by 0.272. The framework integrates longitudinal prediction, missing-modality handling, auxiliary comorbidity modelling and subgroup evaluation within a common pipeline.
\end{abstract}

\begin{graphicalabstract}
\centering
\includegraphics[
  width=\textwidth,
  height=0.75\textheight,
  keepaspectratio
]{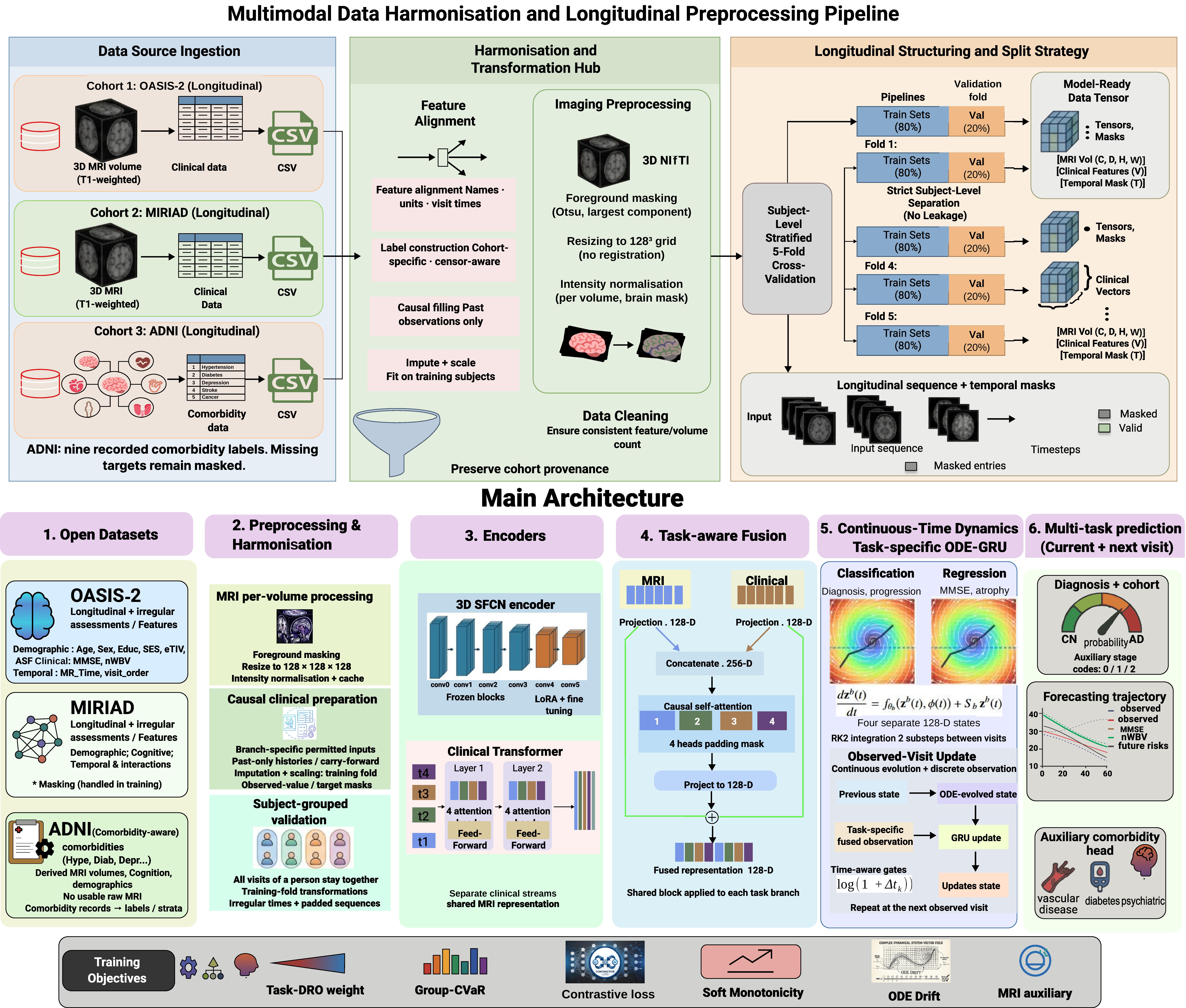}

\vspace{6pt}
\begin{minipage}{1\textwidth}
\small\justifying
\textbf{Graphical abstract.} Top: multi-cohort harmonisation and leakage-controlled preprocessing with subject-grouped five-fold cross-validation. Bottom: SFCN--ODE-GRU architecture jointly predicting diagnosis, progression, MMSE, atrophy and comorbidities, trained with Task-DRO and Group-CVaR.
\end{minipage}
\end{graphicalabstract}

\begin{highlights}
\item Task-specific input controls support multimodal longitudinal AD prediction.
\item ODE-GRU dynamics model irregular visits across multiple prediction tasks.
\item Three-seed internal diagnosis and progression AUROCs reach 0.935 and 0.884.
\item Comorbidity-aware training integrates auxiliary supervision and subgroup auditing.
\item OASIS-3 hybrid diagnosis and next-visit progression AUROCs are 0.767 and 0.764.
\end{highlights}

\end{frontmatter}

\section{Introduction}

Alzheimer's disease (AD) is a progressive neurodegenerative disorder and a major cause of dementia. Global dementia burden and its associated societal costs motivate improved diagnosis and prognosis \cite{gbd2022dementia,wimo2023worldwide}. These estimates concern dementia overall and should not be attributed exclusively to AD. Reliable forecasts of progression may support patient management, clinical trial design and therapeutic research.

However, AD progression is highly heterogeneous, influenced by genetic factors, lifestyle and comorbidities such as vascular disease, diabetes and depression. Lamar et al. \cite{lamar2009vascular} demonstrated that vascular comorbidities significantly affect qualitative error patterns in executive function tests among AD patients, suggesting that vascular pathology modifies the cognitive presentation of AD rather than simply accelerating progression. Using latent Dirichlet allocation, Rajamaki et al. \cite{rajamaki2024comorbidity} identified distinct comorbidity patterns in individuals with Alzheimer's disease compared to those without, revealing that certain comorbidities co-occur more frequently in AD patients. Duara and Barker \cite{duara2022heterogeneity} reviewed heterogeneity in AD diagnosis and progression rates, emphasising implications for therapeutic trials. Maciejewska et al. \cite{maciejewska2021review} provided a comprehensive review of the mechanisms underlying selected comorbidities in Alzheimer's disease. Moreover, clinical visits occur at irregular intervals, and data modalities (e.g., structural MRI, cognitive tests) are often missing. These challenges render many existing machine learning models impractical for real-world clinical decision support. As highlighted by Li et al. \cite{li2026clis}, building models for longitudinal clinical data faces additional hurdles, including mismatched dimensionality between high-dimensional images and low-dimensional tabular variables, as well as inconsistent intervals in follow-up data.

Existing work addresses subsets of these challenges: irregular sampling, shared prediction tasks, comorbidity information and subgroup performance. The proposed model addresses these challenges jointly.

\paragraph{Irregular sampling and discrete-time assumption}
Some forecasting methods place irregular observations on a fixed grid or require missing-value handling \cite{nguyen2020predicting}. Continuous-time neural models provide an alternative representation of elapsed time \cite{chen2018neural,rubanova2019latent,kidger2020neural,fang2025neural}; the experimental protocol therefore compares an ODE formulation with a time-aware discrete GRU.

\paragraph{Disjoint task formulation and static weighting}
Diagnosis (current visit status) and prognosis (future cognitive decline) are typically treated separately, missing the opportunity to learn shared representations that benefit both tasks. Multitask learning has been applied sporadically, but often with fixed, heuristic task weights that ignore varying task difficulty, label noise and data availability \cite{kendall2018multi, chen2018gradnorm, zhou2011multi}. Multitask learning has shown promise in clinical settings \cite{harutyunyan2019multitask}. Nevertheless, these approaches do not address comorbidity interactions.

\paragraph{Neglect of comorbidities and subgroup fairness}
Comorbidities can influence the clinical presentation and progression of AD \cite{lamar2009vascular,duara2022heterogeneity,adrisk2025}. Subgroup-aware learning and distributionally robust optimisation provide tools for examining uneven performance \cite{sagawa2019distributionally,pfohl2021comparison,wang2024using}. Recorded-comorbidity strata inform the robust training objective and the subgroup evaluation.

\paragraph{Temporal leakage and unverifiable explanations}
Longitudinal AD benchmarks are especially exposed to information leakage: visits of one subject placed in both training and test sets, features that are imputed backwards from future visits, cognitive scores that are themselves part of the diagnostic definition, or preprocessing statistics computed on the whole cohort \cite{wen2020convolutional}. Such leaks inflate reported performance and make post-hoc explanations unreliable, because a model that has seen the answer has no reason to rely on disease biology. Few studies test for leakage empirically, and fewer still relate their explanations to known clinical and biological markers of AD.

To address these gaps, this study introduces a \textbf{unified, leakage-controlled framework} that integrates 3D structural MRI and longitudinal clinical data with continuous-time dynamics, multitask learning and explicit comorbidity modelling. The main contributions are:

\begin{enumerate}
    \item \textbf{Multimodal multitask model for irregular visits.} An SFCN brain-age encoder, whose last two blocks are adapted by fine-tuning, low-rank adaptation (LoRA) and an image-level objective, is fused with four task-specific causal clinical Transformers through shared causal attention. Each branch is evolved over irregular visit times with an ODE-GRU. A single model jointly predicts current diagnosis, \textbf{3-year stage-1 progression}, current and next-visit MMSE, next-visit atrophy, and nine comorbidities, with task balancing by a scale-free, capped Task-DRO.

    \item \textbf{Comorbidity-aware learning and subgroup audit.} Nine comorbidities derived from ADNI medical histories are predicted as an auxiliary task under an explicit availability mask; a Group-CVaR penalty targets the worst-performing cohort and comorbidity strata during training; and both primary tasks are evaluated within every comorbidity stratum.

    \item \textbf{Partition controls and empirical leakage audits.} All visits of a subject stay in one fold; all inputs are computed causally (last observation carried forward only, rates from past visits only); preprocessing statistics are fitted on the training fold; branch-specific prohibited inputs prevent circular use of diagnostic criteria; and three empirical checks (partition, feature causality, model causality) are run for every fold model.

    \item \textbf{Task-specific explanations.} Integrated Gradients and group occlusion describe model dependence on available clinical and biological inputs. The explanations are interpreted in the light of the branch-specific input rules.

    \item \textbf{Systematic evaluation.} The evaluation comprises three training seeds, eight paired component and modality ablations with subject-level bootstrap tests, calibration analysis, a per-stratum subgroup audit and external validation on the held-out OASIS-3 cohort.
\end{enumerate}

The remainder of this paper is organised as follows. Section~\ref{sec:related} reviews related work in multimodal AD diagnosis, continuous-time modelling, multitask learning and robust optimisation. Section~\ref{sec:method} describes the proposed architecture and training objectives. Section~\ref{sec:experiments} details the datasets, preprocessing pipeline, leakage controls and experimental setup. Section~\ref{sec:results} presents the main results, ablations, subgroup analyses and interpretability analyses. Section~\ref{sec:discussion} discusses the implications and limitations, and Section~\ref{sec:conclusion} concludes the paper.

\section{Related Work}
\label{sec:related}

\subsection{Multimodal Diagnosis and Prognosis}
Reviews describe machine-learning and deep-learning approaches to AD diagnosis and conversion prediction \cite{gruesoMachineLearningMethods2021,afifi2025vision,sun2026review}. Residual imaging networks, joint prediction of clinical outcomes, multimodal recurrent networks, latent representations and contrastive fusion illustrate distinct modelling strategies \cite{abrolDeepResidualLearning2020,zhangPredictingFutureClinical2012,leePredictingAlzheimersDisease2019,zhou2019latent,multimodalContrastive2023,agostinho2024predicting}. Recent work includes explainable multimodal prediction, longitudinal survival models, clinical--MRI embeddings and cognitive-decline prediction \cite{simkhada2026multimodal,zhang2026longitudinal,lu2026clinical,geng2026machine,kiani2026deep,li2026clis}. Their outcomes, horizons and validation protocols differ, so published scores are not treated as matched benchmarks here. B\'elanger et al. studied dementia identification in home-health assessment records \cite{belanger2023identifying}; their OASIS assessment system is distinct from the OASIS neuroimaging cohorts used in this study.

\subsection{Continuous-Time Modelling}
\label{sec:continuous}
Fixed-grid forecasting has been used in the TADPOLE setting \cite{nguyen2020predicting,marinescu2019tadpole}. Neural ODEs, latent ODEs, GRU-ODE models and neural controlled differential equations provide representations for irregularly sampled sequences \cite{chen2018neural,rubanova2019latent,debrouwer2019gru,kidger2020neural,fang2025neural}. Applications include longitudinal volumetric modelling and multidimensional AD progression models \cite{lachinov2024learning,bossa2023multidimensional}. Sharma et al.'s cLDM-ODE, continuous-time biomarker operators, progression-guided MRI generation and stage-aware mixture-of-experts models address related forecasting problems \cite{zhang2026cldm,wang2026learning,zhao2026forecasting,he2025stage}.

\subsection{Multitask Learning and Task Balancing}
Uncertainty weighting, GradNorm and clinical time-series benchmarks motivate shared representations and adaptive task weights \cite{kendall2018multi,chen2018gradnorm,harutyunyan2019multitask}. AD applications include longitudinal progression, correlated cognitive outcomes and task-relationship learning \cite{zhou2011multi,fan2024multi,wang2019multi,liang2022exploiting,ma2026predicting,zhou2026modeling}. Survival analysis provides a complementary formulation \cite{jahani2025assessing}. The proposed Task-DRO update adapts the weights to relative losses on top of fixed base weights.

\subsection{Comorbidities and Robust Learning}
\label{sec:robust}
Clinical studies and reviews examine vascular effects, comorbidity patterns and AD heterogeneity \cite{lamar2009vascular,rajamaki2024comorbidity,duara2022heterogeneity,maciejewska2021review,adrisk2025}. Progression subtypes and longitudinal survival models address patient heterogeneity from complementary perspectives \cite{ren2025utilizing,karaman2025deep}. Group-DRO, healthcare subgroup comparisons and under-represented subgroup fine-tuning motivate robust objectives \cite{sagawa2019distributionally,pfohl2021comparison,wang2024using}. Site bias, demographic alignment and population-specific dementia models also highlight transportability concerns \cite{wachinger2021detecting,ho2026advancing,liang2026population}. The Group-CVaR penalty targets cohort and recorded-comorbidity strata; its effect on the worst strata is assessed directly.

\subsection{Comparative Summary}
Table~\ref{tab:comparison} compares selected studies by publication year, modality coverage, continuous-time modelling, diagnosis/prognosis, comorbidity modelling, robustness objective, task and reported result. Entries provide context and do not establish a matched performance ranking.
\begin{table*}[!t]
\centering
\scriptsize
\setlength{\tabcolsep}{3pt}
\caption{Comparison of related Alzheimer's disease progression models: capabilities and reported results.}
\label{tab:comparison}
\renewcommand{\arraystretch}{1.08}

\begin{tabularx}{\textwidth}{@{}>{\RaggedRight\arraybackslash}p{0.22\textwidth}cccccc>{\RaggedRight\arraybackslash}X>{\RaggedRight\arraybackslash}p{0.14\textwidth}@{}}
\toprule
\textbf{Method} & \textbf{Year} & \textbf{M} & \textbf{CT} & \textbf{D+P} & \textbf{Com} & \textbf{R} & \textbf{Task (metric)} & \textbf{Reported} \\
\midrule
\multicolumn{9}{c}{\textit{Multimodal Diagnosis and Conversion}} \\
\midrule
Zhang \& Shen \cite{zhangPredictingFutureClinical2012} & 2012 & \checkmark & $\times$ & $\circ$ & $\times$ & $\times$ & MCI conv.\ (AUC) & 0.77$^\dagger$ \\
Lee et al. \cite{leePredictingAlzheimersDisease2019} & 2019 & \checkmark & $\times$ & $\circ$ & $\times$ & $\times$ & MCI$\to$AD conv.\ (AUC) & 0.86 \\
Zhou et al. \cite{zhou2019latent} & 2019 & \checkmark & $\times$ & \checkmark & $\times$ & $\times$ & sMCI/pMCI (AUC) & 0.755 \\
Huang \cite{multimodalContrastive2023} & 2023 & \checkmark & $\times$ & $\circ$ & $\times$ & $\times$ & Diagnosis, ADNI (Acc) & 0.838 \\
Afifi et al. \cite{afifi2025vision} & 2026 & -- & -- & -- & -- & -- & Review & -- \\
Simkhada et al. \cite{simkhada2026multimodal} & 2026 & \checkmark & $\times$ & $\circ$ & $\times$ & $\times$ & sMCI/pMCI, 2 y (BA) & 0.820 \\
Lu et al. \cite{lu2026clinical} & 2026 & \checkmark & $\times$ & $\circ$ & $\times$ & $\times$ & Progression, 4 y (AUC) & 0.91--0.92 \\
Geng \& Zhang \cite{geng2026machine} & 2026 & \checkmark & $\times$ & $\circ$ & $\times$ & $\times$ & MMSE decline, 12 mo (AUC) & 0.773 \\
Zhang et al. \cite{zhang2026longitudinal} & 2026 & \checkmark & $\circ$ & $\circ$ & $\times$ & $\times$ & MCI$\to$AD, 1 y (cdAUC) & 0.909 \\
\midrule
\multicolumn{9}{c}{\textit{Longitudinal and Continuous-Time Modelling}} \\
\midrule
Nguyen et al. \cite{nguyen2020predicting} & 2020 & \checkmark & $\times$ & \checkmark & $\times$ & $\times$ & Dx (mAUC) / ADAS-Cog13 (MAE) & 0.944 / 4.30 \\
Rubanova et al. \cite{rubanova2019latent} & 2019 & $\times$ & \checkmark & -- & -- & $\times$ & Non-AD (method) & -- \\
Bossa \& Sahli \cite{bossa2023multidimensional} & 2023 & $\circ$ & \checkmark & \checkmark & $\times$ & $\times$ & MCI$\to$AD, 3 y (AUC) & 0.86 \\
Sharma et al. (cLDM-ODE) \cite{zhang2026cldm} & 2026 & \checkmark & \checkmark & $\circ$ & $\times$ & $\times$ & AD conv.\ (Acc) / MMSE (MAE) & 0.88 / 1.5 \\
He et al. (stage-aware MoE) \cite{he2025stage} & 2026 & $\times$ & \checkmark & $\circ$ & $\times$ & $\times$ & Regional tau forecast (\(r\)) & 0.717 \\
Wang et al. \cite{wang2026learning} & 2026 & \checkmark & \checkmark & $\circ$ & $\times$ & $\times$ & Biomarker maps (rel.\ \(L^2\) acc.) & $\approx$0.90 \\
Zhao et al. \cite{zhao2026forecasting} & 2026 & $\times$ & $\times$ & $\circ$ & $\times$ & $\times$ & 4-y MRI forecast (SSIM) & 0.931 \\
\midrule
\multicolumn{9}{c}{\textit{Multitask Learning}} \\
\midrule
Kendall et al. \cite{kendall2018multi} & 2018 & $\times$ & $\times$ & -- & -- & $\times$ & Non-AD (method) & -- \\
Chen et al. \cite{chen2018gradnorm} & 2018 & $\times$ & $\times$ & -- & -- & $\times$ & Non-AD (method) & -- \\
Zhou et al. \cite{zhou2011multi} & 2011 & \checkmark & $\times$ & $\circ$ & $\times$ & $\times$ & Future MMSE/ADAS-Cog & n.r. \\
Wang et al. \cite{wang2019multi} & 2019 & $\times$ & $\times$ & $\circ$ & $\times$ & $\times$ & Future MMSE / ADAS-Cog (\(r\)) & 0.625 / 0.664 \\
Liang et al. \cite{liang2022exploiting} & 2023 & $\times$ & $\times$ & $\circ$ & $\times$ & $\times$ & 18 cognitive scores (wR) & 0.416 \\
Jahani et al. \cite{jahani2025assessing} & 2025 & \checkmark & $\circ$ & $\circ$ & $\times$ & $\times$ & MCI$\to$AD (C-index) & 0.878 \\
Ma et al. \cite{ma2026predicting} & 2026 & $\circ$ & $\times$ & \checkmark & $\times$ & $\times$ & Dx (Acc) / ADAS-Cog (\(R^2\)) & 0.928 / 0.82 \\
Zhou et al. \cite{zhou2026modeling} & 2026 & $\times$ & $\times$ & $\times$ & $\times$ & $\times$ & 7 phenotypes, ADNI (\(r\)) & 0.158 \\
\midrule
\multicolumn{9}{c}{\textit{Comorbidity, Heterogeneity \& Robust Learning}} \\
\midrule
Lamar et al. \cite{lamar2009vascular} & 2010 & -- & -- & -- & \checkmark & -- & Clinical study & -- \\
Maciejewska et al. \cite{maciejewska2021review} & 2021 & -- & -- & -- & \checkmark & -- & Review & -- \\
Duara \& Barker \cite{duara2022heterogeneity} & 2022 & -- & -- & -- & \checkmark & -- & Review & -- \\
Wachinger et al. \cite{wachinger2021detecting} & 2021 & $\times$ & $\times$ & $\times$ & $\times$ & $\circ$ & Site-bias detection (Acc) & 0.715 \\
Ren et al. \cite{ren2025utilizing} & 2026 & \checkmark & $\times$ & $\circ$ & $\times$ & $\times$ & Trajectory, 48 mo (Acc) & 0.791 \\
Karaman et al. \cite{karaman2025deep} & 2026 & \checkmark & $\circ$ & $\circ$ & $\times$ & $\times$ & Future AD dx (AUROC) & n.r. \\
Wang et al. \cite{wang2024using} & 2025 & $\circ$ & $\times$ & $\circ$ & $\times$ & \checkmark & Dementia accuracy-gap reduction & $>$3\% \\
Ho et al. \cite{ho2026advancing} & 2026 & $\times$ & $\times$ & $\circ$ & $\times$ & \checkmark & Ethnic BA gap (pp) & 6.9$\to$1.4 \\
Liang et al. \cite{liang2026population} & 2026 & $\circ$ & $\times$ & $\circ$ & $\times$ & $\circ$ & Population-specific risk (AUC) & 0.835--0.903 \\
\midrule
\multicolumn{9}{c}{\textit{This Study}} \\
\midrule
\textbf{Proposed framework} & \textbf{2026} & \checkmark & \checkmark & \checkmark & \checkmark & \checkmark & \makecell{3-y MCI conv.\ / sMCI-pMCI\\ / dx (AUC)} & \makecell{\textbf{0.884} / \textbf{0.870}\\ / \textbf{0.935}} \\
\bottomrule
\end{tabularx}
\vspace{2pt}
\noindent\footnotesize M: multimodal; CT: continuous-time; D+P: diagnosis and prognosis; Com: explicit comorbidity modelling; R: robust optimisation or bias mitigation. $\checkmark$: addressed; $\circ$: partially addressed; $\times$: not addressed in the cited work; --: not applicable. Results differ in cohorts, endpoints, horizons, metrics and validation protocols and are not head-to-head comparisons. n.r.: no single summary metric reported. Values for the proposed model are three-seed internal-validation means.
\end{table*}

The proposed architecture combines causal clinical encoders, an adapted MRI encoder, task-specific ODE-GRU branches, auxiliary recorded-comorbidity prediction and subgroup objectives. The component and modality ablations assess this integrated design.

\section{Methodology}
\label{sec:method}

\begin{figure*}[t]
\centering
\includegraphics[width=\textwidth]{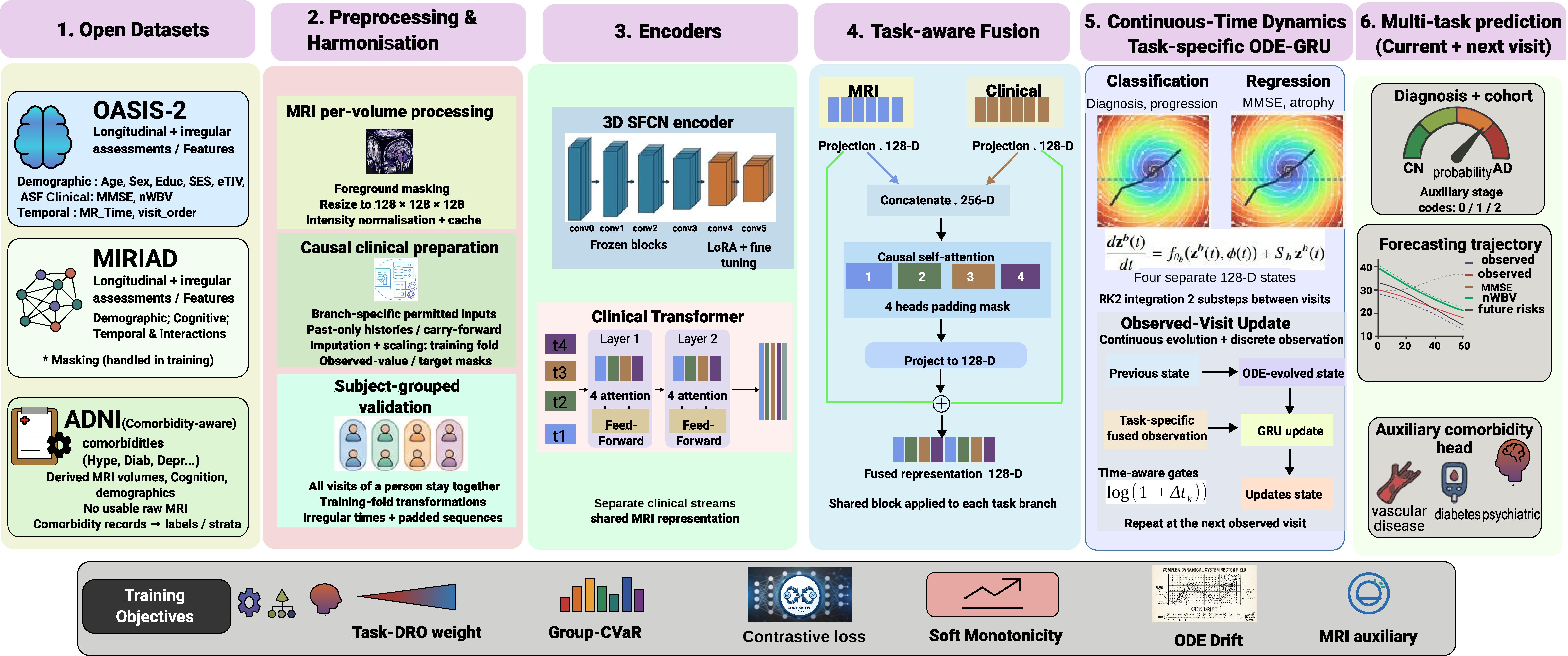}
\caption{SFCN--ODE-GRU architecture. A LoRA-adapted SFCN MRI encoder and four task-specific causal clinical Transformers are combined by shared causal fusion; each task branch evolves an ODE-GRU state between irregular visits and feeds the diagnosis, progression, MMSE, atrophy and comorbidity heads. Icons are illustrative.}
\label{fig:pipeline}
\end{figure*}

Figure~\ref{fig:pipeline} provides an overview of the proposed framework. For each subject \(i\) the model processes a sequence of visits \(\{(x^{\text{MRI}}_{i,k}, \mathbf{x}^{\text{Clin}}_{i,k}, t_{i,k})\}_{k=1}^{T_i}\), where \(t_{i,k}\) is the time in years since the first visit. It outputs, at every visit, the current diagnosis, the probability of conversion to a more advanced stage within three years, current and next-visit MMSE, next-visit atrophy, and, at the last visit, a multi-label comorbidity vector. The visit-level prediction paths are causal: the output at visit \(k\) depends only on visits \(1,\dots,k\).

\subsection{Per-Visit Encoding}

\subsubsection{MRI Encoder}
The MRI encoder uses the Simple Fully Convolutional Network (SFCN) pretrained for brain-age estimation on UK Biobank \cite{peng2021accurate}. The first four convolutional blocks are frozen (including BatchNorm statistics), and the 64-dimensional pooled output is projected to a latent vector \(\mathbf{h}^M_{i,k}\in\mathbb{R}^{128}\). The last two blocks (conv4, conv5; 1.79M parameters) are fine-tuned with a small learning rate, and their convolutions are additionally wrapped with LoRA adapters \cite{hu2022lora}:
\begin{equation}
\mathrm{Conv}'(\mathbf{x}) = \mathrm{Conv}(\mathbf{x}) + \tfrac{\alpha}{r}\, B\bigl(A(\mathbf{x})\bigr),
\label{eq:lora}
\end{equation}
where \(A\) is a rank-\(r\) convolution with the original kernel size, \(B\) is a \(1\times1\times1\) convolution initialised at zero (so that training starts exactly at the pretrained encoder), \(r=8\) and \(\alpha=16\). Because brain-age pretraining does not directly optimise the cohort-defined diagnosis endpoint, an auxiliary cohort-defined binary diagnosis head (\(128\to64\to2\)) is applied to \(\mathbf{h}^M_{i,k}\) alone on visits with MRI and trained with cross-entropy (weight \(\lambda_{\text{img}}=0.2\)), which encourages diagnosis-relevant image representations. Visits without MRI use a learnable embedding \(\mathbf{e}_{\text{noMRI}}\), and a learned availability embedding is added to every MRI latent. During training the whole MRI stream of a subject is replaced by \(\mathbf{e}_{\text{noMRI}}\) with probability 0.05 (modality dropout), so that the model remains usable when imaging is absent. SFCN is preferred to larger 3D CNNs because it is pretrained on brain MRI and has about 3.0M parameters (46.6M for a 3D ResNet-50).

\subsubsection{Task-Specific Causal Clinical Encoders}
Clinical inputs differ by task, because a feature that is legitimate for one task can be circular for another (Section~\ref{sec:leakage}). The architecture therefore includes four clinical encoders with identical architecture but different input sets: \emph{diagnosis}, \emph{forecast}, \emph{MMSE} and \emph{atrophy}. Each concatenates its time-varying features with temporal features (time since first visit and visit order), projects them to 128 dimensions and applies a 2-layer, 4-head Transformer encoder with a causal mask and a key-padding mask:
\begin{equation}
\mathbf{h}^{C,b}_{i,1:T_i} = \mathrm{Transformer}_b\bigl(W_b[\mathbf{x}^{b}_{i,1:T_i}; \mathbf{x}^{\text{time}}_{i,1:T_i}]\bigr),\quad b\in\mathcal{B},
\end{equation}
with \(\mathcal{B}=\{\text{diag},\text{fc},\text{mmse},\text{atr}\}\). The causal mask guarantees that visit \(k\) never attends to visits \(k'>k\).

\subsection{Shared Causal Attention Fusion}
One fusion module, shared by all four task branches, combines the MRI latent with each task-specific clinical stream. The two representations are projected separately to 128 dimensions and concatenated to form a 256-dimensional sequence:
\begin{equation}
\begin{aligned}
\mathbf{m}_{i,k}&=P_M\mathbf{h}^{M}_{i,k},\qquad
\mathbf{c}^{b}_{i,k}=P_C\mathbf{h}^{C,b}_{i,k},\\
\mathbf{H}^{b}_{i,1:T_i}&=[\mathbf{m}_{i,1:T_i};\mathbf{c}^{b}_{i,1:T_i}].
\end{aligned}
\end{equation}
Four-head self-attention uses this same concatenated sequence for queries, keys and values, with causal and padding masks. The attended output is projected to 128 dimensions, passed through dropout, and combined with a scaled modality residual:
\begin{equation}
\begin{aligned}
\mathbf{A}^{b}_{i,1:T_i}
&=\mathrm{MHA}_{\mathrm{causal,pad}}
 (\mathbf{H}^{b}_{i,1:T_i},\mathbf{H}^{b}_{i,1:T_i},\mathbf{H}^{b}_{i,1:T_i}),\\
\mathbf{F}^{b}_{i,1:T_i}
&=\mathrm{Dropout}(P_O\mathbf{A}^{b}_{i,1:T_i})\\
&\quad+\lambda_g\left(w_M\mathbf{m}_{i,1:T_i}+w_C\mathbf{c}^{b}_{i,1:T_i}\right),\\
\mathbf{u}^{b}_{i,k}&=\mathrm{Adapt}_b(\mathbf{F}^{b}_{i,k}),
\end{aligned}
\label{eq:fusion}
\end{equation}
where $(w_M,w_C)=\mathrm{softmax}(\mathbf{a})$ are learned global modality weights and $\lambda_g=0.01$. These weights are shared across subjects, visits and tasks; they are not causal modality shares or patient-specific explanations. Each $\mathrm{Adapt}_b$ is a separate two-layer MLP with a SiLU activation. The shared fusion module is called separately for the diagnosis, progression, MMSE and atrophy clinical branches.

\subsection{Continuous-Time Dynamics: ODE-GRU}
\label{sec:ode}

Each branch has its own ODE-GRU cell \cite{rubanova2019latent, debrouwer2019gru} with a 128-dimensional hidden state \(\mathbf{z}^b(t)\). \emph{Between visits}, the state evolves according to a learned vector field:
\begin{equation}
\frac{d\mathbf{z}^b(t)}{dt} = f_{\theta_b}\bigl(\mathbf{z}^b(t), \phi(t)\bigr) + S_b\,\mathbf{z}^b(t),
\label{eq:ode}
\end{equation}
where \(f_{\theta_b}\) is a 3-layer MLP (256 hidden units, SiLU) whose last layer is initialised at zero, \(S_b\) is a small orthogonally initialised linear skip, and \(\phi(t)\) is a sinusoidal time embedding (8 frequencies from \(1/8\) to \(16\) rad/year). Equation~\eqref{eq:ode} is integrated from \(t_{i,k-1}\) to \(t_{i,k}\) with the midpoint (RK2) method using two substeps per visit interval under the evaluated configuration. \emph{At a visit}, a GRU update mixes the propagated state with the fused observation \(\mathbf{u}^b_{i,k}\), using update and reset gates that also receive the elapsed time:
\begin{equation}
\begin{aligned}
\mathbf{g}_k &= \sigma\bigl(W_g[\mathbf{z}^{b-}_{k};\mathbf{u}^b_{k};\log(1+\Delta t_k)]\bigr),\quad \mathbf{g}\in\{\mathbf{z}_g,\mathbf{r}_g\},\\
\tilde{\mathbf{n}}_k &= \tanh\bigl(W_n[\mathbf{r}_g\odot\mathbf{z}^{b-}_{k};\mathbf{u}^b_{k}]\bigr),\\
\mathbf{z}^{b}_{k} &= (1-\mathbf{z}_g)\odot\mathbf{z}^{b-}_{k} + \mathbf{z}_g\odot\tilde{\mathbf{n}}_k,
\end{aligned}
\label{eq:odegru}
\end{equation}
where \(\mathbf{z}^{b-}_{k}\) is the ODE-propagated state just before visit \(k\) and \(\Delta t_k=t_{i,k}-t_{i,k-1}\). The evolved sequence is refined by a causal temporal self-attention layer with a residual connection and layer normalisation. Compared with a discrete GRU, this formulation gives the latent disease state a well-defined value \emph{at any time between visits}. An ODE-drift penalty \(\mathcal{L}_{\text{drift}}=\mathbb{E}\|f_{\theta_b}(\mathbf{z},\phi(t))+S_b\mathbf{z}\|^2\) (\(\lambda_{\text{drift}}=10^{-3}\)) discourages unnecessarily large vector fields.

\subsection{Prediction Heads}
\label{sec:heads}

Classification and next-visit residual heads use evolved branch states; current MMSE and volume heads instead use the task-adapted fused representations \(\mathbf{u}^b_{i,k}\):

\begin{itemize}
    \item \textbf{Current diagnosis} (cohort-defined binary endpoint): \(\hat{y}^{\text{dx}}_{i,k} = \mathrm{softmax}(\mathrm{MLP}_{\text{dx}}(\mathbf{z}^{\text{diag}}_{i,k}))\). An auxiliary 3-way stage head (0/1/2; cohort-specific mapping) on the same state uses all visits including MCI (weight 0.01).
    \item \textbf{3-year conversion}: \(\hat{y}^{\text{conv}}_{i,k} = \mathrm{softmax}\bigl(\mathrm{MLP}_{\text{conv}}(W_m\mathbf{z}^{\text{fc}}_{i,k} + W_s\mathbf{s}_{i})\bigr)\), where \(\mathbf{s}_i\) denotes the covariate vector used by the direct head pathway at visit \(k\) (demographics, baseline imaging and fluid biomarkers, creatinine and family history, with family history treated as a distinct predictor from the participant's own medical history) entering through a linear projection. Each variable enters with its causally available value at visit \(k\) (last observation carried forward; variables recorded only at baseline enter as recorded). The vector is a subset of the diagnosis-branch inputs and therefore obeys the same prohibited-input rules.
    \item \textbf{Current MMSE and atrophy}: \(\hat{m}_{i,k}=\mathrm{MLP}_{\text{mmse}}(\mathbf{u}^{\text{mmse}}_{i,k})\), \(\hat{a}_{i,k}=\mathrm{MLP}_{\text{atr}}(\mathbf{u}^{\text{atr}}_{i,k})\). The atrophy target is normalised whole-brain volume (nWBV) in OASIS-2 and intracranial-volume-normalised hippocampal volume in ADNI, standardised per cohort on the training fold.
    \item \textbf{Next-visit MMSE and atrophy} (residual forecasts): \(\hat{m}^{\text{next}}_{i,k} = m^{\text{obs}}_{i,k} + \Delta\hat{m}_{i,k}\), where \(m^{\text{obs}}_{i,k}\) is the current observed value (or the model's current estimate when unobserved) and \(\Delta\hat{m}_{i,k}=\mathrm{MLP}_{\Delta}(\mathbf{z}^{\text{mmse}}_{i,k})\); atrophy is handled identically. The residual head learns a change relative to the current value. The observed current MMSE (or volume) is an explicit forecast anchor, even though that variable is excluded from the corresponding current-estimation encoder. The head is not conditioned on the interval to the next visit; next-visit results therefore describe prediction at the next observed visit, whose interval varies, rather than at a fixed horizon.
    \item \textbf{Comorbidities} (multi-label, 9 outputs): \[\hat{\mathbf{c}}_i = \sigma(\mathrm{MLP}_{\text{com}}(\mathbf{z}^{\text{diag}}_{i,T_i}))\] read from the diagnosis-branch state at the last visit, so that comorbidity information shapes the representation used for diagnosis.
\end{itemize}

\subsection{Robust Optimisation Objective}
The training objective combines task balancing, subgroup robustness and trajectory regularisation.

\subsubsection{Base Task Losses}
Each task \(\tau\) uses a masked loss
\begin{equation}
\mathcal{L}_\tau = \frac{\sum_i \sum_k m^\tau_{i,k} \,\ell_\tau(i,k)}{\sum_i \sum_k m^\tau_{i,k} + \epsilon},
\label{eq:masked_loss}
\end{equation}
where \(m^\tau_{i,k}\) indicates that the label is \emph{actually observed} at that visit (imputed values are never used as targets). \(\ell_\tau\) is cross-entropy with label smoothing (0.1) for diagnosis, focal loss for conversion (from epoch 5), Huber loss (\(\beta=2\)) for MMSE and atrophy, and binary cross-entropy with capped positive-class weights for comorbidities. The task set is
\begin{equation*}
\begin{aligned}
\mathcal{T} = \{&\text{dx},\ \text{dx}_3,\ \text{conv},\ \text{MMSE}_{\text{cur+next}},\\
&\text{atrophy}_{\text{cur+next}},\ \text{com}\}.
\end{aligned}
\end{equation*}

\subsubsection{Comorbidity Mask}
Rule-derived recorded-comorbidity information is available only for ADNI participants. For OASIS-2 and MIRIAD, comorbidities are treated as \textit{unknown}, not as zeros. A subject-level availability mask distinguishes observed records from unavailable information:
\[
m_i^{\text{com}} = \begin{cases}
1 & \text{ADNI medical history available},\\
0 & \text{otherwise}.
\end{cases}
\]
The comorbidity loss, the comorbidity part of the Group-CVaR penalty and all comorbidity subgroup analyses use only subjects with \(m_i^{\text{com}}=1\). This prevents conflating ``no comorbidity'' with ``missing data''.

\subsubsection{Task-DRO: Scale-Free Dynamic Task Balancing}
Task losses lie on very different scales (focal vs.\ Huber vs.\ BCE), so a softmax over raw losses would be dominated by whichever loss is numerically largest. The entropy-regularised DRO update \cite{sagawa2019distributionally} is therefore applied to \emph{relative} losses. Each task keeps an exponential moving average \(\bar{\mathcal{L}}_\tau\) (decay 0.98), and
\begin{equation}
q_\tau = \mathrm{CappedSoftmax}_{q_{\max}}\!\left(\frac{\log\bigl(1+\mathcal{L}_\tau/\bar{\mathcal{L}}_\tau\bigr)}{\gamma}\right),
\label{eq:softmax_update}
\end{equation}
with \(\gamma=0.5\). The capped softmax iteratively clips any weight above \(q_{\max}=0.3\) and redistributes the remaining mass, so that no single task can dominate. The primary loss is \(\mathcal{L}_{\text{TaskDRO}} = |\mathcal{T}|\sum_\tau q_\tau w_\tau \mathcal{L}_\tau\), where \(w_\tau\) are base weights ramped in by a smooth curriculum (diagnosis from epoch 1, MMSE/atrophy and comorbidity ramping in during the first epochs). The base weights are \(w_{\text{dx}}=0.5\), \(w_{\text{dx3}}=0.01\), \(w_{\text{conv}}=1.5\), \(w_{\text{MMSE}}=0.1\), \(w_{\text{atr}}=0.05\) and \(w_{\text{com}}=0.2\). A task with a larger loss relative to its recent average receives a larger adaptive weight.

\subsubsection{Group-CVaR over Cohorts and Comorbidity Strata}
To target groups with higher losses, the objective includes a Conditional Value-at-Risk penalty \cite{sagawa2019distributionally} on per-subject classification losses (diagnosis and conversion). For a set of groups \(\mathcal{G}\) present in a mini-batch, with mean loss \(\mathcal{L}_g\),
\begin{equation}
\mathcal{R}_{\mathcal{G}} = \eta + \frac{1}{(1-\alpha)|\mathcal{G}|} \sum_{g \in \mathcal{G}} \max(0, \mathcal{L}_g - \eta),
\label{eq:cvar}
\end{equation}
where \(\eta\) is the \(\alpha\)-quantile (linear interpolation) of the group mean losses, \(\alpha=0.7\). The penalty is applied twice: over cohorts (ADNI, OASIS-2, MIRIAD) and over coarse comorbidity strata \{none, cardiovascular (hypertension, CAD, stroke), metabolic (diabetes, hyperlipidaemia), depression, other (arthritis, cancer, COPD)\} for subjects with \(m^{\text{com}}_i=1\). The strata are mutually exclusive: a subject with several conditions is assigned by priority depression \(>\) cardiovascular \(>\) metabolic \(>\) other. The penalty is skipped when fewer than two groups or fewer than two subjects per group are present. The penalty is computed from unreduced per-subject losses, so that group means genuinely differ. \(\mathcal{R}_{\text{CVaR}} = \mathcal{R}_{\text{cohort}} + \mathcal{R}_{\text{com}}\), with \(\lambda_{\text{cvar}}=0.1\).

\subsubsection{Monotonicity Constraints}
\label{sec:monotonicity}
For subjects with a dementia label at any visit, a small hinge penalty \(\mathcal{L}_{\text{mono}}\) (\(\lambda_{\text{mono}}=0.01\)) discourages increases in predicted MMSE and atrophy-target values between consecutive visits, averaged over visit pairs. Here the atrophy targets are volume measures (nWBV or normalised hippocampal volume), so the discouraged direction is an increase in volume, rather than an increase in tissue loss. The soft penalty applies to the current-value estimates of MMSE and the atrophy target. It encodes a preference for non-increasing trajectories rather than a hard biological constraint: upward predictions remain possible, and actual measurements may fluctuate. Its purpose is regularisation, not a claim that every individual follows a strictly monotonic course.

\subsubsection{Longitudinal Contrastive Regularisation}
To encourage subject-specific and temporally consistent diagnosis representations, a small MLP projects the evolved states at the last two visits of each subject, \((\mathbf{z}^{\text{diag}}_{i,T_i-1},\mathbf{z}^{\text{diag}}_{i,T_i})\), and a symmetric NT-Xent loss is applied across the mini-batch:
\begin{equation}
\begin{split}
\mathcal{L}_{\text{con}} = -\tfrac{1}{2B}\sum_{i}\Bigl[&\log\tfrac{e^{s_{ii}/\kappa}}{\sum_j e^{s_{ij}/\kappa}} \\
&+ \log\tfrac{e^{s_{ii}/\kappa}}{\sum_j e^{s_{ji}/\kappa}}\Bigr],
\end{split}
\label{eq:contrastive}
\end{equation}
where \(s_{ij}\) is the cosine similarity between the projected states of subjects \(i\) and \(j\), \(\kappa=0.1\) and \(\lambda_{\text{con}}=0.05\).

\subsubsection{Components Deliberately Disabled}
Uncertainty weighting \cite{kendall2018multi} is not used, to avoid introducing a second task-weighting mechanism alongside Task-DRO.

\subsubsection{Final Objective}
\begin{equation}
\boxed{
\begin{aligned}
\mathcal{L}_{\text{total}} &= \mathcal{L}_{\text{TaskDRO}} + \lambda_{\text{cvar}}\mathcal{R}_{\text{CVaR}} + \lambda_{\text{img}}\mathcal{L}_{\text{img}} + \lambda_{\text{mono}}\mathcal{L}_{\text{mono}} \\
&\quad + \lambda_{\text{con}}\mathcal{L}_{\text{con}} + \lambda_{\text{drift}}\mathcal{L}_{\text{drift}} + \lambda_{\Delta}\|\Delta\hat{\mathbf{y}}\|^2,
\end{aligned}
}
\label{eq:final}
\end{equation}
where \(\mathcal{L}_{\text{img}}\) is the image-level auxiliary loss and the last term (\(\lambda_\Delta=0.05\)) penalises residual forecast magnitude after the first five anchor epochs.

\subsection{Network--Tabular Hybrid and Calibration}
\label{sec:hybrid}
Gradient-boosted trees remain very strong on tabular clinical data. The complementary tabular stack is fitted on the training fold only and comprises the mean rank of a class-balanced logistic regression and a histogram gradient-boosting classifier using the same causal features as the corresponding network branch (no MRI). The hybrid score is a rank mixture \(w\,\mathrm{rank}(\hat{p}_{\text{net}}) + (1-w)\,\mathrm{rank}(\hat{p}_{\text{tab}})\), with one weight \(w\in\{0,0.1,\dots,1\}\) per cohort selected on validation predictions. Probabilities are calibrated by temperature scaling \cite{guo2017calibration}. Within each fold, the temperature is fitted on that fold's validation predictions, and internal calibration metrics are computed on the same predictions; internal calibration is therefore in-sample with respect to the temperature. For external evaluation, the temperatures fitted on the internal folds are applied unchanged to OASIS-3, so external calibration is out-of-sample. Unless stated otherwise, all claims about the model refer to the \emph{network alone}; the hybrid is reported as a complementary result.

\subsection{Training Algorithm}

Algorithm~\ref{alg:train} summarises the training procedure.

\begin{algorithm}[t]
\caption{Training with Task-DRO, Group-CVaR and Auxiliary Terms}
\label{alg:train}
\begin{algorithmic}[1]
\STATE \textbf{Input:} model \(\theta\), tasks \(\mathcal{T}\), cohorts and comorbidity strata \(\mathcal{G}\), \(\gamma\), \(q_{\max}\), \(\alpha\), \(\lambda_{\text{cvar}}\), \(\lambda_{\text{mono}}\), \(\lambda_{\text{con}}\), \(\lambda_{\text{drift}}\).
\FOR{each epoch \(e\)}
    \FOR{each mini-batch of subjects}
        \STATE Encode MRI (SFCN + LoRA) and the four clinical streams (causal Transformers)
        \STATE Fuse each stream with MRI (Eq.~\ref{eq:fusion}); evolve with ODE-GRU (Eqs.~\ref{eq:ode}--\ref{eq:odegru})
        \STATE Compute masked task losses \(\mathcal{L}_\tau\) on observed labels only (Eq.~\ref{eq:masked_loss})
        \STATE Update EMA \(\bar{\mathcal{L}}_\tau\); compute capped Task-DRO weights \(q_\tau\) (Eq.~\ref{eq:softmax_update})
        \STATE \(\mathcal{L}_{\text{TaskDRO}} \leftarrow |\mathcal{T}|\sum_\tau q_\tau w_\tau(e)\mathcal{L}_\tau\)
        \STATE \textbf{// Group-CVaR on unreduced per-subject losses}
        \STATE \(\mathcal{R}_{\text{CVaR}} \leftarrow \mathcal{R}_{\text{cohort}} + \mathcal{R}_{\text{com}}\) (Eq.~\ref{eq:cvar}; groups with \(<2\) subjects skipped; \(\mathcal{R}_{\text{com}}\) only on \(m^{\text{com}}=1\))
        \STATE Compute \(\mathcal{L}_{\text{img}}\) on visits with MRI, \(\mathcal{L}_{\text{mono}}\) (Sec.~\ref{sec:monotonicity}), \(\mathcal{L}_{\text{con}}\) (Eq.~\ref{eq:contrastive}), and \(\mathcal{L}_{\text{drift}}\)
        \STATE Compute \(\mathcal{L}_{\text{total}}\) (Eq.~\ref{eq:final}); accumulate gradients (2 steps), clip, AdamW update
    \ENDFOR
    \STATE Evaluate on the validation fold; keep the checkpoint with the best selection score; early stopping
\ENDFOR
\STATE Fit temperature and conversion threshold on validation predictions; fit the tabular stack on the training fold
\end{algorithmic}
\end{algorithm}

\section{Experimental Setup}
\label{sec:experiments}

\subsection{Datasets}
\label{sec:data}

The development dataset combines three publicly available longitudinal cohorts with heterogeneous populations, imaging protocols and comorbidity profiles: OASIS-2 \cite{oasis_portal}, MIRIAD \cite{miriad} and ADNI \cite{adni_portal}. The assembled ADNI subset contains 2,430 subjects. The aggregated development set comprises \textbf{2,649 subjects and 17,317 visits}, with \textbf{1,081 indexed MRI acquisitions across 896 MRI-available visits from 219 subjects} (Table~\ref{tab:dataset_stats}). OASIS-3 is held out entirely for external testing.

\begin{table}[t]
\centering
\caption{Dataset statistics per source. MIRIAD provides several acquisitions per visit (708 acquisitions over 523 visits).}
\label{tab:dataset_stats}
\begin{tabular}{lccc}
\toprule
Source & Subjects & Visits & MRI acquisitions \\
\midrule
OASIS-2 & 150 & 373 & 373 \\
MIRIAD & 69 & 523 & 708 \\
ADNI & 2,430 & 16,421 & 0 \\
\midrule
\textbf{Total} & \textbf{2,649} & \textbf{17,317} & \textbf{1,081} \\
\bottomrule
\end{tabular}
\end{table}

\textbf{Diagnosis label mapping.} The primary task is a \emph{cohort-defined binary diagnosis endpoint}, not a uniform CN-versus-dementia definition. ADNI excludes MCI visits from this task, MIRIAD contrasts controls and AD, and OASIS-2 contrasts CDR $<1$ with CDR $\ge1$. In particular, OASIS-2 CDR 0.5 visits belong to the negative class and are not cognitively normal controls. The class counts are 4,526 negative and 2,839 positive visits. Table~\ref{tab:diagnosis_mapping} gives the mapping; stage labels are harmonisation codes rather than interchangeable clinical diagnoses.

\begin{table}[t]
\centering
\caption{Diagnosis label mapping per cohort. Stage codes: 0 = CN/control/CDR 0, 1 = ADNI MCI or OASIS-2 CDR 0.5, 2 = dementia/AD/CDR $\ge1$. Stage 1 is not clinically equivalent across cohorts.}
\label{tab:diagnosis_mapping}
\small
\setlength{\tabcolsep}{2pt}
\begin{tabular}{lll}
\toprule
Cohort & Source labels & Binary label (stage) \\
\midrule
OASIS-2 & CDR 0 / 0.5 / \(\ge\)1 & \makecell[l]{0 = CDR \(<\)1 (0, 1)\\ 1 = CDR \(\ge\)1 (2)} \\
MIRIAD & Control, AD & \makecell[l]{0 = control (0)\\ 1 = AD (2)} \\
ADNI & CN, MCI, Dementia & \makecell[l]{0 = CN (0)\\ 1 = Dementia (2)\\ MCI: excluded (1)} \\
\bottomrule
\end{tabular}
\normalsize
\end{table}

\textbf{Nominal 3-year progression label.} For visits with stage code 0 or 1, a positive label is assigned when the first later observed visit with a higher stage occurs within 3 years (tolerance 0.15 years, i.e.\ \(\le3.15\) years); a negative label requires no observed increase and at least 2.85 years of follow-up. Shorter follow-up without progression is treated as an unknown outcome. Stage decreases are not events: conversion is defined by the first observed stage increase. There are 5,375 labelled at-risk visits, including 1,443 positives (ADNI 27\%, OASIS-2 23\%). MIRIAD has no stage-1 at-risk visits. Evaluation covers three views: (i) MCI visits, (ii) first-MCI-visit sMCI versus pMCI, and (iii) all stage-0/1 visits. For ADNI, MCI-to-dementia is a clinical conversion endpoint; OASIS-2 CDR-stage progression is different. The MCI views contain stage-1 visits from ADNI (clinical MCI) and OASIS-2 (CDR 0.5): 3,005 visits and 850 subjects.

\subsubsection{OASIS-2}
OASIS-2 provides MRI, age, sex, education, SES, eTIV, nWBV, ASF and MMSE; it lacks PET, CSF, regional volumes, other cognitive tests and medical history. CDR defines the label and is never an input.

\subsubsection{MIRIAD}
MIRIAD provides repeated T1 scans, age, sex and MMSE; it lacks education, SES, brain volumes, PET, CSF and medical history. The label is the study group (control or AD); CDR is not used as an input.

\subsubsection{ADNI}
ADNI provides regional MRI volumes \cite{fischl2012freesurfer}, PET, CSF, APOE4, cognitive and functional tests, creatinine and family history; raw ADNI MRI is not used. Medications, the participant's own medical-history items and condition-defining laboratory values (glucose, HbA\(_{1c}\), cholesterol, triglycerides) are excluded from the inputs because they define the comorbidity labels; family history is a separate predictor and is not used to construct those labels.

\subsubsection{Comorbidity Representation}
Nine binary comorbidities (hypertension, diabetes, depression, hyperlipidaemia, arthritis, cancer, coronary artery disease (CAD), stroke and COPD) are derived from ADNI medical histories by keyword rules that ignore negated and family-history mentions. They are available for 1,739 ADNI subjects (\(m_i^{\text{com}}=1\)), from 835 hypertension to 54 COPD cases; for OASIS-2 and MIRIAD they are unknown, not zero.

\subsection{Preprocessing and Leakage Control}

\begin{figure*}[t]
\centering
\includegraphics[width=\textwidth]{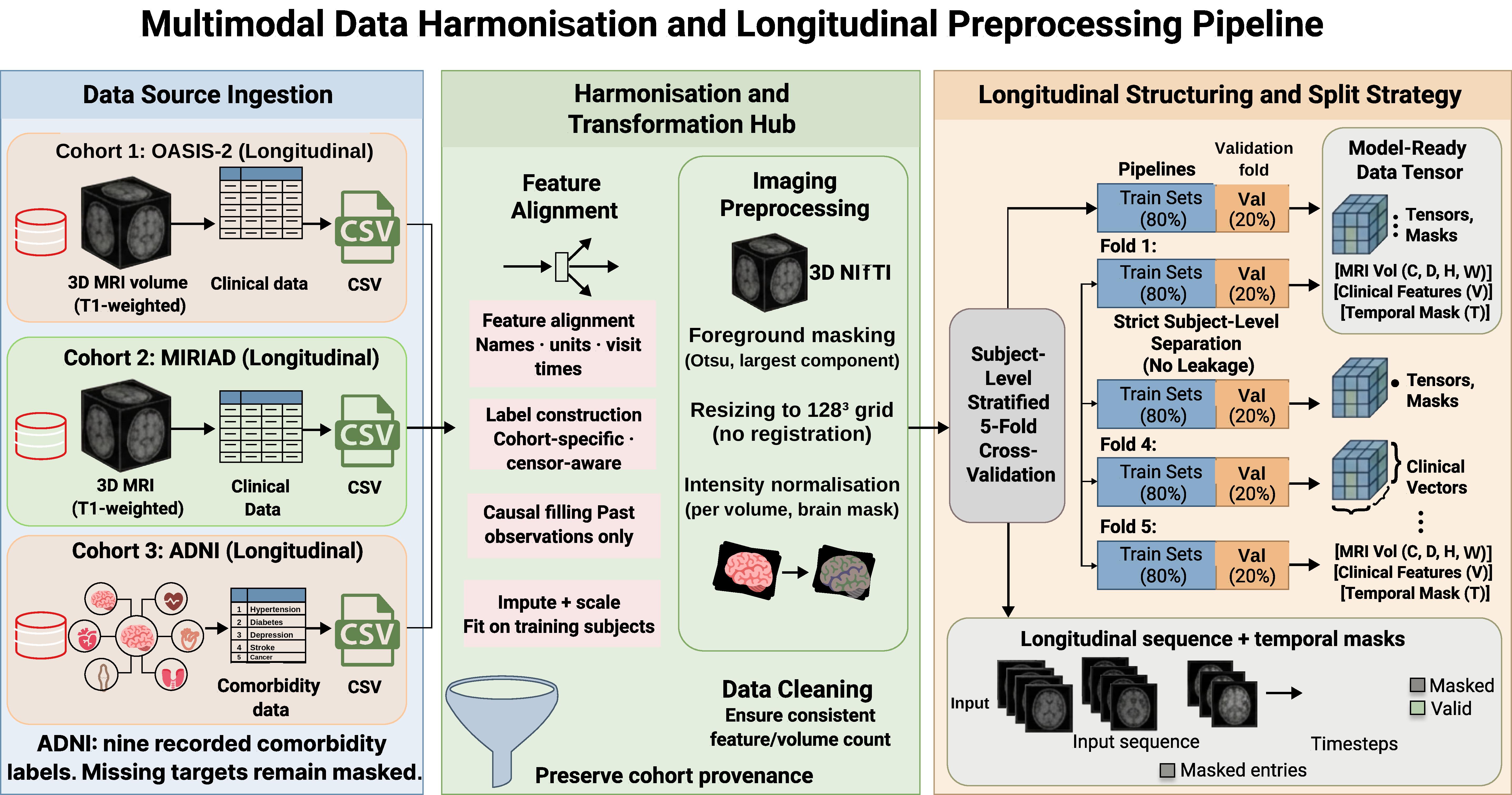}
\caption{Data preparation and internal validation. Cohorts are aligned, labels constructed and features filled causally; imputation and scaling are fitted on training subjects only, MRI is processed per volume, and all visits of a participant remain within one of five subject-grouped folds.}
\label{fig:preprocessing}
\end{figure*}

Figure~\ref{fig:preprocessing} summarises cohort alignment, subject-level folds, fold-wise transformation fitting and model-ready sequences.

\label{sec:leakage}

\subsubsection{Imaging}
T1 volumes undergo foreground masking using Otsu thresholding and the largest connected component, are resized to \(128^3\) and intensity-normalised with the statistics of the brain mask of the \emph{same volume}. No dataset-level statistic is involved. No spatial registration or voxel-spacing resampling is applied: each volume is resized as a whole to the fixed grid, and the encoder operates in native orientation.

\subsubsection{Causal Feature Construction}
Time is measured in years since each subject's first visit. Sparse longitudinal measures (MMSE, volumes, PET, CSF, cognitive tests) are filled with \emph{last observation carried forward only}; the first visit is never back-filled, and each filled value is accompanied by an availability flag and the time since its last observation. Rates of change are computed from the two most recent observations \emph{at or before} the current visit (minimum gap 0.25 years). All remaining medians, scalers, target standardisers and the comorbidity class weights are fitted on the training fold and applied unchanged to the validation fold. Inputs are clipped to \(\pm5\) standard deviations.

\subsubsection{Branch-Specific Prohibited Inputs}
Several variables are part of how the target itself is defined and are therefore removed from the corresponding branch (Table~\ref{tab:prohibited}). In particular, the \emph{diagnosis} branch never sees MMSE, any cognitive or functional test, CDR or the current clinical stage, because these assessments can contribute to clinical diagnosis or encode closely related target information. The diagnosis branch therefore uses biological measures, demographics and history. The forecast branch may use cognitive tests and, for ADNI, the current stage (MCI or not), because these are known at the time of the forecast and are not the future outcome. The MMSE and atrophy branches start from the forecast input set and remove their own targets: the MMSE branch excludes MMSE, its rates and composite scores that contain MMSE, and the atrophy branch excludes nWBV and hippocampal volume and their rates. Under the evaluated configuration, both therefore also receive current clinical-stage information for ADNI (MCI or not), which is part of the clinical diagnostic assessment, together with the remaining cognitive tests and regional volumes. The stage indicator is set only for cohorts with a clinical diagnosis and is never derived from CDR, so OASIS-2 stage is not an input. CDR is excluded everywhere because it defines the OASIS-2 labels. Prohibited inputs are removed when each branch's input list is built, and an assertion checks each branch against its prohibited list.

\begin{table}[t]
\centering
\caption{Branch-specific prohibited inputs (removed when input lists are built).}
\label{tab:prohibited}
\scriptsize
\setlength{\tabcolsep}{3pt}
\begin{tabular}{lp{0.74\columnwidth}}
\toprule
Branch & Prohibited inputs \\
\midrule
Diagnosis & CDR, MMSE, all cognitive/functional tests, current stage \\
Forecast & CDR (and its sum of boxes) \\
MMSE & CDR, MMSE and its rates, composite scores that contain MMSE \\
Atrophy & CDR, nWBV, hippocampal volume and their rates \\
\bottomrule
\end{tabular}
\end{table}

\subsubsection{Cross-Validation}
Internal evaluation uses 5-fold \texttt{StratifiedGroupKFold} grouped by subject (split seed 42): every visit and every MRI file of a subject belongs to exactly one fold. In each fold the model is trained on four-fifths of the subjects and evaluated on the held-out fifth, which also drives early stopping, the conversion threshold, the temperature and the hybrid weights. The resulting metrics are therefore \emph{internal-validation} estimates. The subject-bootstrap intervals condition on the saved predictions. Mini-batches are balanced across cohorts.

\subsubsection{Empirical Leakage Checks}
Three checks are run for every fold model and written to a fold-level report: \textbf{(P1) partition}: zero subjects and zero MRI files are shared between training and validation, and no label-like column or prohibited input reaches any branch; \textbf{(P2) feature causality}: for 60 random subjects, every derived feature at visit \(k\) is recomputed after deleting all visits after \(k\) and must be identical; \textbf{(P3) model causality}: for 40 random subjects, every model output at visit \(\le k\) must be unchanged (tolerance \(10^{-3}\)) when visits \(>k\) are deleted.

\subsection{Training Details}
Training uses AdamW (weight decay \(5\times10^{-4}\)) with differentiated learning rates: Transformers and fusion \(5\times10^{-5}\), classification heads \(1\times10^{-4}\), regression heads \(3\times10^{-4}\), MMSE head \(3\times10^{-5}\), ODE-GRU \(1\times10^{-4}\), LoRA adapters \(1\times10^{-4}\) and the fine-tuned SFCN blocks \(1\times10^{-5}\). The batch size is 8 subjects with gradient accumulation over 2 steps, automatic mixed precision (ODE integration in float32), gradient-norm clipping at 0.5 and a ReduceLROnPlateau scheduler. Training runs for at most 25 epochs with early stopping (patience 5, minimum 7 epochs) on a validation selection score \(0.35\,\text{AUC}_{\text{dx}} + 0.45\,\text{AUC}_{\text{conv}} + 0.10\,S_{\text{MMSE}} + 0.10\,\text{AUC}_{\text{com}}\), where \(S_{\text{MMSE}}=\mathrm{clip}(1-\text{MAE}_{\text{next}}/\text{MAE}_{\text{persist}},0,1)\) compares the next-visit MMSE error with that of carrying the current value forward. The latent dimension is 128 throughout.

The remaining hyperparameters are \(\gamma=0.5\), \(q_{\max}=0.3\), \(\alpha=0.7\), \(\lambda_{\text{cvar}}=0.1\), \(\lambda_{\text{mono}}=0.01\), \(\lambda_{\text{con}}=0.05\) (\(\kappa=0.1\)), \(\lambda_{\text{drift}}=10^{-3}\), \(\lambda_{\text{img}}=0.2\), LoRA \(r=8\), \(\alpha_{\text{LoRA}}=16\).

The full model is trained with three seeds (42, 123, 2026) on the same folds. Training uses one NVIDIA A40 GPU per run.

\subsection{Evaluation Metrics}
Primary metrics use pooled out-of-fold predictions: the corresponding network was not fitted on the subject, but its fold was reused for model selection and post-processing. 95\% confidence intervals use a subject-level bootstrap (1,000 resamples). NR: not reported. Unless stated otherwise, \(\pm\) denotes the standard deviation across the three training seeds; fold-level standard deviations are labelled as such. Arrows indicate the optimisation direction.

\begin{itemize}
    \item \textbf{Diagnosis:} AUROC \(\uparrow\), average precision \(\uparrow\), balanced accuracy \(\uparrow\), expected calibration error (ECE) \(\downarrow\) before and after temperature scaling, with 10 equal-width bins; bins with fewer than five predictions are omitted.
    \item \textbf{3-year conversion:} AUROC \(\uparrow\) on the three views defined above; conversion ECE uses all eligible CN+MCI visits rather than MCI visits alone.
    \item \textbf{Cognition and atrophy:} current and next-visit MMSE MAE \(\downarrow\) and \(R^2\) \(\uparrow\); next-visit atrophy \(R^2\) \(\uparrow\). Regression metrics use only \emph{truly observed} targets, never imputed values.
    \item \textbf{Comorbidities:} per-condition AUROC \(\uparrow\) and mean AUROC over the nine conditions (ADNI subjects with \(m^{\text{com}}=1\)).
    \item \textbf{Subgroup robustness:} diagnosis and conversion AUROC within each of the nine comorbidity strata (subjects \emph{with} the condition); each task is summarised by the worst stratum (the lowest value, never the best), the mean of the three worst, and the disparity (best minus worst).
    \item \textbf{Paired comparisons:} \(\Delta\)AUROC between two models on the same subjects, with a 95\% CI from a paired subject-level bootstrap (1,000 resamples); a CI excluding zero is described as nominal evidence of a difference. No multiple-comparison adjustment is applied.
\end{itemize}

\subsection{Ablation Protocol}
The ablations reuse the same five folds and seed 42, so comparisons can be paired at the subject level. Most remove one component; the comorbidity variant jointly removes its prediction task and its CVaR strata:
\begin{enumerate}
    \item \textbf{w/o Task-DRO:} fixed curriculum weights \(w_\tau\).
    \item \textbf{w/o Group-CVaR:} \(\lambda_{\text{cvar}}=0\).
    \item \textbf{w/o contrastive:} \(\lambda_{\text{con}}=0\).
    \item \textbf{w/o comorbidity head:} no comorbidity task and no comorbidity strata in CVaR.
    \item \textbf{w/o LoRA:} LoRA adapters removed (conv4--5 still fine-tuned).
    \item \textbf{Clinical-only:} MRI replaced by the no-MRI embedding for all subjects.
    \item \textbf{Imaging-focused:} clinical encoder inputs removed and temporal features retained. The diagnosis head then sees only the image embedding and visit-time features, whereas the conversion head still receives the static-covariate projection, so its conversion results are not image-only.
    \item \textbf{GRU (no ODE):} identical cell and time-aware gates (Eq.~\ref{eq:odegru}), but the state is held constant between visits instead of being integrated.
\end{enumerate}

\section{Results}
\label{sec:results}

\subsection{Main Results}

\begin{figure*}[t]
\centering
\includegraphics[width=0.92\textwidth]{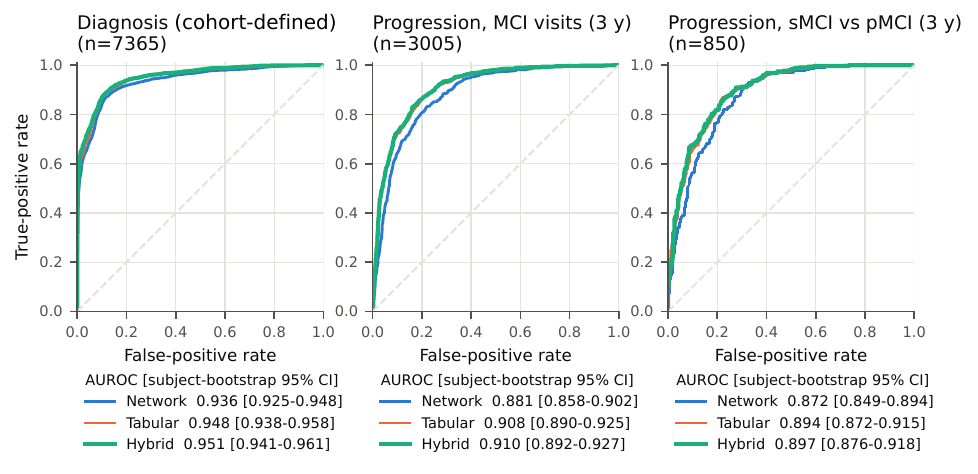}
\caption{ROC curves on pooled out-of-fold predictions (seed 42) for diagnosis and two conversion views (all MCI visits; first MCI visit per subject), for the network, the tabular stack and the hybrid. Legends give the AUROC with its 95\% subject-bootstrap CI. The two progression panels use different visit-selection rules and denominators; they are complementary evaluation views, not independent replication.}
\label{fig:roc}
\end{figure*}

Table~\ref{tab:main} summarises the full model on pooled out-of-fold predictions. The network reaches a diagnosis AUROC of \textbf{0.936} [0.925, 0.948] (0.935\(\pm\)0.002 over three seeds) and a \textbf{3-year stage-1 progression AUROC of 0.881} [0.858, 0.902] (0.884\(\pm\)0.003). At the subject level, the first stage-1 visit separates stable from progressive cases with an AUROC of 0.872 (0.870\(\pm\)0.005). Combining the network with the tabular stack gives \textbf{0.951}, \textbf{0.909} and \textbf{0.896} (three-seed means). The tabular stack alone reaches 0.948, 0.908 and 0.894 (seed 42; Figure~\ref{fig:roc}), so the hybrid improves only slightly on it. Current MMSE is estimated with an MAE of 1.58 points and next-visit MMSE with an MAE of 1.62 points (\(R^2=0.692\)). Temperature scaling lowers the diagnosis ECE from 0.135 to 0.031. Across comorbidity strata, diagnosis stays at or above \textbf{0.819} (worst: COPD) and conversion at or above \textbf{0.765} (worst: stroke). The auxiliary comorbidity head achieves a mean AUROC of 0.565 from the permitted inputs. Seed standard deviations are below 0.03 for all listed metrics. Figure~\ref{fig:roc} shows the ROC curves.

\begin{table}[t]
\caption{Overall performance of the full model (SFCN, internal 5-fold validation). Value: seed 42, pooled out-of-fold, with 95\% subject-bootstrap CI; last column: mean\(\pm\)SD over seeds 42/123/2026. Hybrid = network + tabular stack.}
\label{tab:main}
\centering
\scriptsize
\setlength{\tabcolsep}{3pt}
\resizebox{\columnwidth}{!}{%
\begin{tabular}{lcc}
\toprule
Metric & Value [95\% CI] & 3 seeds \\
\midrule
\multicolumn{3}{l}{\textit{Diagnosis (cohort-defined binary endpoint)}} \\
AUROC, network \(\uparrow\) & 0.936 [0.925, 0.948] & 0.935\(\pm\)0.002 \\
AUROC, hybrid \(\uparrow\) & 0.951 [0.941, 0.961] & 0.951\(\pm\)0.001 \\
Average precision, network / hybrid \(\uparrow\) & 0.917 / 0.936 & 0.915 / 0.936 \\
Balanced accuracy \(\uparrow\) & 0.870 & 0.865\(\pm\)0.008 \\
ECE, raw \(\to\) calibrated \(\downarrow\) & 0.135 \(\to\) 0.031 & 0.134 \(\to\) 0.025 \\
\midrule
\multicolumn{3}{l}{\textit{3-year conversion}} \\
MCI visits, network \(\uparrow\) & 0.881 [0.858, 0.902] & 0.884\(\pm\)0.003 \\
MCI visits, hybrid \(\uparrow\) & 0.910 & 0.909\(\pm\)0.001 \\
sMCI vs.\ pMCI, network \(\uparrow\) & 0.872 [0.849, 0.894] & 0.870\(\pm\)0.005 \\
sMCI vs.\ pMCI, hybrid \(\uparrow\) & 0.897 & 0.896\(\pm\)0.001 \\
CN+MCI visits, network / hybrid \(\uparrow\) & 0.858 / 0.888 & 0.858 / 0.888 \\
ECE, raw \(\to\) calibrated \(\downarrow\) & 0.167 \(\to\) 0.057 & 0.178 \(\to\) 0.059 \\
\midrule
\multicolumn{3}{l}{\textit{Cognition and atrophy}} \\
Current MMSE MAE \(\downarrow\) & 1.577 & 1.592\(\pm\)0.021 \\
Next-visit MMSE MAE \(\downarrow\) & 1.615 & 1.611\(\pm\)0.010 \\
Next-visit MMSE \(R^2\) \(\uparrow\) & 0.692 & 0.698\(\pm\)0.006 \\
Next-visit atrophy \(R^2\) \(\uparrow\) & 0.828 & 0.826\(\pm\)0.002 \\
\midrule
\multicolumn{3}{l}{\textit{Comorbidity and subgroups (ADNI, \(m^{\text{com}}=1\))}} \\
Comorbidity mean AUROC \(\uparrow\) & 0.565 & 0.571\(\pm\)0.010 \\
Worst-stratum diagnosis \(\uparrow\) & 0.819 (COPD) & 0.827\(\pm\)0.008 (COPD) \\
Worst-stratum conversion \(\uparrow\) & 0.765 (stroke) & 0.774\(\pm\)0.029 \\
Hardest comorbidity \(\uparrow\) & 0.512 (diabetes) & 0.517\(\pm\)0.006 \\
\bottomrule
\end{tabular}%
}
\par\footnotesize Entries without brackets: no CI computed. Seed SD is distinct from a bootstrap CI. Worst-stratum conversion is the lowest stratum in each seed (stroke, diabetes or COPD).
\end{table}

\subsection{External Validation on OASIS-3}
\label{sec:external}

OASIS-3 was excluded from model development and reserved for external validation
after the model, fold preprocessors, MMSE and atrophy scalers, tabular baselines
and temperature scalers were frozen on the internal folds. The five fold models
trained on ADNI, OASIS-2 and MIRIAD were applied unchanged to every OASIS-3
visit; no OASIS-3 label, image or statistic entered fitting, threshold selection,
calibration or hybrid weighting. An attribute-based overlap screen was applied to the OASIS-3 candidate pool before model evaluation to reduce potential participant overlap with OASIS-2. A subject was flagged
when sex, education and handedness matched an OASIS-2 participant, the OASIS-3
baseline age exceeded the OASIS-2 baseline age by 3--20 years, the absolute
difference in estimated total intracranial volume was at most
30{,}000~mm\textsuperscript{3}, and the OASIS-3 baseline age was at least 60.
The screen excluded 135 participants; the frozen exclusion list is reused by the
build script, which asserts that no excluded subject appears in the prepared
cohort. One T1 was indexed per MRI session, and each session was linked to the
nearest UDS visit within $\pm270$~days, a matching window that permits imaging and clinical assessments to occur months apart. The prepared cohort contains 1{,}152 subjects and 2{,}407 visits
(1{,}995 diagnosis eligible, 707 progression eligible, 980 comorbidity eligible);
every prepared visit carries a T1, so the MRI encoder was evaluated on all
records. Within each seed, every fold model is applied with its own training-fold
preprocessing, tabular stack and hybrid weight; each fold's saved temperature is applied to
its own predictions, and raw and calibrated predictions are then averaged per subject--visit
across the five fold models, with the decision threshold set to the mean of the fold
thresholds. The primary external result is the per-visit average of the three
seeds' predictions, computed before any metric is evaluated. External progression is evaluated at the next observed visit, not at a fixed 3-year horizon. The external analysis evaluates next-visit transfer separately from the internal 3-year endpoint.

Table~\ref{tab:external} summarises the external results. The diagnosis AUROC is 0.763 for the network and 0.767 for the hybrid. For next-observed-visit progression, the network achieves 0.733 and the hybrid 0.764, compared with 0.710 for logistic regression. The tabular stack achieves a progression AUROC of 0.768, while logistic regression achieves the highest diagnosis point estimate (0.787). Transferred temperature scaling reduces diagnosis ECE from 0.197 to 0.052; progression ECE changes from 0.401 to 0.303. Next-visit MMSE prediction achieves an MAE of 0.86 and $R^2=0.24$, and current MMSE estimation achieves an MAE of 1.80 and $R^2=-0.03$. Comorbidity prediction was evaluated externally for eight of the nine conditions (hypertension, diabetes, depression, hyperlipidaemia, arthritis, cancer, coronary artery disease and stroke); COPD was not evaluable because none of the 980 OASIS-3 subjects with comorbidity records was COPD-positive. The mean comorbidity AUROC across these eight conditions is 0.565. These results extend the evaluation to a separate cohort across diagnostic, progression, cognitive and auxiliary outcomes. External atrophy prediction is outside the present evaluation.

\begin{table*}[t]
\centering
\small
\caption{External validation on OASIS-3, three-seed per-visit ensemble. Counts
denote eligible records; diagnosis and progression are AUROC; MMSE rows report
MAE\,/\,$R^2$. Progression uses the next observed visit rather than the internal 3-year horizon. No paired confidence intervals are available for these contrasts.}
\label{tab:external}
\begin{tabular}{lcccc}
\toprule
Task / metric & Network & Tabular & Hybrid & Logistic \\
\midrule
Diagnosis AUROC ($n=1{,}995$) & 0.763 & 0.762 & 0.767 & 0.787 \\
Progression AUROC ($n=707$) & 0.733 & 0.768 & 0.764 & 0.710 \\
Current MMSE MAE\,/\,$R^2$ ($n=2{,}407$) & 1.80\,/\,$-0.03$ & -- & -- & -- \\
Next-visit MMSE MAE\,/\,$R^2$ ($n=1{,}255$) & 0.86\,/\,0.24 & -- & -- & -- \\
Diagnosis ECE, raw $\to$ calibrated & 0.197 $\to$ 0.052 & -- & -- & -- \\
Progression ECE, raw $\to$ calibrated & 0.401 $\to$ 0.303 & -- & -- & -- \\
\bottomrule
\end{tabular}
\par\vspace{3pt}\footnotesize Mean comorbidity AUROC: 0.565 ($n=980$). The
evaluation includes fewer conditions than the nine-output training head.
\end{table*}

\subsection{Context with Published Results}

The studies in Table~\ref{tab:comparison} use different cohorts, label definitions, horizons and validation designs, so their scores provide context rather than a ranking.

\subsection{Ablation Studies}

Table~\ref{tab:ablation_detailed} reports all ablations together with paired subject-bootstrap tests against the full model on the same subjects. For the component ablations, the observed differences in pooled diagnosis, MCI-visit conversion and sMCI/pMCI AUROC lie between \(-0.0038\) and \(+0.0036\), and the paired intervals include zero except for MCI-visit conversion without Group-CVaR (\(+0.0036\) [0.0001, 0.0069], nominal and unadjusted). Key observations:

\begin{itemize}
    \item \textbf{Modalities.} Removing the clinical encoder inputs (imaging-focused) lowers diagnosis by 0.272 [0.252, 0.293], MCI-visit conversion by 0.051 [0.031, 0.072] and sMCI/pMCI by 0.056 [0.033, 0.078], and raises the MMSE error from 1.58 to 2.83. Removing MRI (clinical-only) gives differences whose intervals include zero (diagnosis \(-0.002\) [\(-0.007\), 0.004]); in this configuration the pooled signal is carried mainly by clinical inputs and derived volumes.
    \item \textbf{Longitudinal contrastive loss.} Without it, the mean comorbidity AUROC falls from 0.565 to 0.540, with lower values for eight of the nine conditions.
    \item \textbf{LoRA.} Without LoRA (conv4--5 still fine-tuned), the worst-stratum diagnosis AUROC is 0.891, compared with 0.819 for the full model; the worst strata are small and these are single-seed estimates.
    \item \textbf{Task-DRO.} Without Task-DRO, the MMSE errors are 1.554 (current) and 1.599 (next visit), compared with 1.577 and 1.615.
    \item \textbf{Comorbidity head.} Removing the auxiliary task gives paired differences of \(+0.001\) [\(-0.003\), 0.006] and \(-0.002\) [\(-0.008\), 0.005] for diagnosis and MCI-visit conversion, respectively.
    \item \textbf{Group-CVaR.} Its subgroup effect is analysed in Section~\ref{sec:subgroups}.
\end{itemize}

\begin{table*}[t]
\centering
\scriptsize
\setlength{\tabcolsep}{4pt}
\caption{Ablation study (network, seed 42, same five folds). * = paired subject-bootstrap 95\% CI of the AUROC difference to the full model excludes 0 (the three AUROC columns). The last row gives the seed standard deviation of the full model (seeds 42/123/2026) to describe variation over three runs, not as a significance threshold. Asterisks denote nominal, unadjusted contrasts.}
\label{tab:ablation_detailed}
\resizebox{\textwidth}{!}{%
\begin{tabular}{lcccccccc}
\toprule
\multirow{2}{*}{Experiment} & Diagnosis & \multicolumn{2}{c}{3-year conversion AUROC \(\uparrow\)} & \multicolumn{3}{c}{Cognition / atrophy} & \multicolumn{2}{c}{Comorbidity (ADNI)} \\
\cmidrule(lr){3-4} \cmidrule(lr){5-7} \cmidrule(lr){8-9}
 & AUROC \(\uparrow\) & MCI visits & sMCI/pMCI & MMSE MAE \(\downarrow\) & Next MMSE MAE \(\downarrow\) & Next atrophy \(R^2\) \(\uparrow\) & Mean AUROC \(\uparrow\) & Worst-stratum dx \(\uparrow\) \\
\midrule
\textbf{Full model} & \textbf{0.936} & \textbf{0.881} & \textbf{0.872} & \textbf{1.577} & \textbf{1.615} & \textbf{0.828} & \textbf{0.565} & \textbf{0.819} \\
\midrule
\multicolumn{9}{c}{\textbf{Component ablations}} \\
\midrule
w/o Task-DRO & 0.937 & 0.880 & 0.869 & 1.554 & 1.599 & 0.833 & 0.577 & 0.834 \\
w/o Group-CVaR & 0.937 & 0.885* & 0.874 & 1.597 & 1.619 & 0.834 & 0.565 & 0.828 \\
w/o contrastive & 0.938 & 0.884 & 0.871 & 1.557 & 1.594 & 0.839 & 0.540 & 0.837 \\
w/o comorbidity head & 0.937 & 0.880 & 0.870 & 1.574 & 1.610 & 0.834 & \NR & \NR \\
w/o LoRA & 0.938 & 0.880 & 0.875 & 1.594 & 1.617 & 0.835 & 0.565 & 0.891 \\
\midrule
\multicolumn{9}{c}{\textbf{Modality ablations}} \\
\midrule
Clinical-only & 0.935 & 0.887 & 0.870 & 1.575 & 1.611 & 0.829 & 0.574 & 0.787 \\
Imaging-focused & 0.664* & 0.830* & 0.817* & 2.825 & 2.233 & 0.582 & \NR & \NR \\
\midrule
\multicolumn{9}{c}{\textbf{Temporal model}} \\
\midrule
GRU, no ODE (time-aware gates) & 0.937 & 0.884 & 0.869 & 1.576 & 1.603 & 0.837 & 0.570 & 0.851 \\
\midrule
\textit{Full model, seed SD} & \textit{\(\pm\)0.002} & \textit{\(\pm\)0.003} & \textit{\(\pm\)0.005} & \textit{\(\pm\)0.021} & \textit{\(\pm\)0.010} & \textit{\(\pm\)0.002} & \textit{\(\pm\)0.010} & \textit{\(\pm\)0.008} \\
\bottomrule
\end{tabular}%
}
\par\vspace{2pt}\footnotesize * Paired subject-bootstrap 95\% CI of the difference to the full model excludes 0 (nominal, unadjusted); markers apply to the three AUROC columns only. NR: comorbidity prediction metrics are unavailable where the auxiliary head is absent; the saved outputs of these runs contain no comorbidity labels, so worst-stratum diagnosis was not computed for them. Seed SD: standard deviation of the full model over seeds 42/123/2026. Paired differences quoted in the text are computed from unrounded AUROCs and can differ by 0.001 from differences of the rounded table values.
\end{table*}

\subsection{MRI Encoder Comparison}
\label{sec:encoder}

To assess the choice of MRI encoder, the SFCN encoder was replaced by a MedicalNet-pretrained 3D ResNet-50 with the same adaptation (last-block fine-tuning and LoRA, rank 8) and all other settings unchanged (seed 42). Table~\ref{tab:encoder} reports accuracy and cost. The three encoder configurations reach comparable AUROC, with differences of at most 0.006 and overlapping 95\% subject-bootstrap CIs (ResNet-50: diagnosis 0.932 [0.920, 0.945], MCI conversion 0.884 [0.862, 0.904]), while SFCN uses about one seventh of the total parameters of ResNet-50 and requires less training time. SFCN with LoRA is therefore retained as the MRI encoder.

\begin{table}[t]
\centering
\caption{MRI encoder comparison (seed 42, internal 5-fold validation). Dx: diagnosis AUROC; MCI: MCI-conversion AUROC; sMCI: sMCI/pMCI AUROC. Parameters are for the full model; s/epoch is mean training time per epoch; 5-fold time is total wall-clock time over the five folds. The two SFCN rows repeat the corresponding AUROCs of Table~\ref{tab:ablation_detailed}; this table adds the ResNet-50 comparison and computational cost.}
\label{tab:encoder}
\resizebox{\columnwidth}{!}{%
\begin{tabular}{lcccccc}
\toprule
Encoder & Dx & MCI & sMCI & Params (total/trainable) & s/epoch & 5-fold time \\
\midrule
SFCN + LoRA (main) & 0.936 & 0.881 & 0.872 & 6.97M / 4.02M & 187 & 5.33 h \\
SFCN w/o LoRA & 0.938 & 0.880 & 0.875 & 6.91M / 3.96M & 217 & 5.32 h \\
ResNet-50 + LoRA & 0.932 & 0.884 & 0.871 & 50.50M / 4.34M & 208 & 6.18 h \\
\bottomrule
\end{tabular}%
}
\end{table}

\subsection{Continuous-Time Dynamics}
\label{sec:ode_results}

The ODE-GRU and the time-aware GRU share the same cell and gates and differ only in between-visit integration. The paired differences (GRU minus ODE) are \(+0.001\) [\(-0.003\), 0.004] for diagnosis, \(+0.003\) [\(-0.003\), 0.009] for MCI-visit conversion and \(-0.003\) [\(-0.010\), 0.004] for sMCI/pMCI; no statistically detectable AUROC difference was observed. The ODE formulation additionally provides a latent state defined between visits.

\subsection{Performance per Cohort}

\begin{figure}[t]
\centering
\includegraphics[width=\columnwidth]{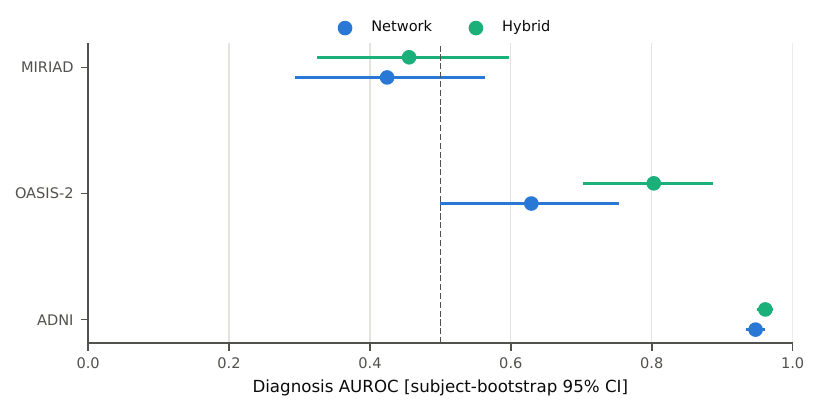}
\caption{Cohort-specific internal diagnosis AUROC for the network and hybrid, seed 42. Intervals use 1,000 subject-cluster bootstrap resamples retaining all visits. Cohort-specific endpoints are heterogeneous. Cohort differences reflect distinct populations, inputs and label mappings; pooled performance should not be interpreted as uniformly strong discrimination in every cohort.}
\label{fig:cohort_diagnosis}
\end{figure}

Table~\ref{tab:cohort} and Figure~\ref{fig:cohort_diagnosis} break down diagnosis by cohort. ADNI, with rich biomarkers, reaches 0.947 (hybrid 0.961), and OASIS-2 reaches 0.629 (hybrid 0.803). On MIRIAD, the diagnosis AUROC of the fused model is 0.424, which is no better than chance: there, the diagnosis branch sees only demographics and the MRI embedding, because MMSE is prohibited, and this cohort has fewer permitted clinical inputs than ADNI. When the clinical encoder inputs are removed (imaging-focused setting), the image pathway reaches 0.597 on MIRIAD and 0.621 on OASIS-2. The imaging-focused comparison complements the fused-model evaluation by characterising performance under reduced clinical input.

\begin{table}[t]
\centering
\caption{Diagnosis AUROC per cohort (seed 42, pooled out-of-fold).}
\label{tab:cohort}
\scriptsize
\begin{tabular}{lccc}
\toprule
Cohort & Visits & Network & Hybrid \\
\midrule
ADNI & 6,469 & 0.947 & 0.961 \\
OASIS-2 & 373 & 0.629 & 0.803 \\
MIRIAD & 523 & 0.424 & 0.456 \\
\bottomrule
\end{tabular}
\end{table}

\subsection{Comorbidity Prediction}

Table~\ref{tab:comorbidity_per_label} reports the per-condition AUROC. Using inputs that exclude medication, medical-history and condition-defining laboratory records, the auxiliary head achieves a mean AUROC of 0.565: coronary artery disease (0.634) and depression (0.605) are the best detected, and diabetes (0.512) and hyperlipidaemia (0.519) are close to chance. Group occlusion of the comorbidity head (not shown) indicates that the remaining signal comes mainly from demographics (up to 0.08 AUROC) and, for COPD, from structural volumes (0.05). Without Group-CVaR, the mean comorbidity AUROC is also 0.565, and without the longitudinal contrastive loss it is 0.540 (Table~\ref{tab:ablation_detailed}).

\begin{table}[t]
\centering
\caption{Per-condition comorbidity AUROC, ADNI subjects with \(m^{\text{com}}=1\) (\(n=1{,}739\)). Full: seed 42; w/o CVaR: ablation, seed 42.}
\label{tab:comorbidity_per_label}
\scriptsize
\setlength{\tabcolsep}{3pt}
\begin{tabular}{lcccc}
\toprule
Condition & Cases & Full & w/o CVaR & 3-seed mean \\
\midrule
Depression & 518 & 0.605 & 0.615 & 0.598 \\
Diabetes & 162 & 0.512 & 0.540 & 0.551 \\
Hypertension & 835 & 0.547 & 0.544 & 0.547 \\
Coronary artery disease & 223 & 0.634 & 0.637 & 0.636 \\
Stroke & 74 & 0.570 & 0.546 & 0.588 \\
Hyperlipidaemia & 801 & 0.519 & 0.523 & 0.521 \\
COPD & 54 & 0.554 & 0.529 & 0.544 \\
Arthritis & 708 & 0.558 & 0.569 & 0.564 \\
Cancer & 512 & 0.585 & 0.586 & 0.592 \\
\midrule
Mean & & \textbf{0.565} & 0.565 & 0.571 \\
\bottomrule
\end{tabular}
\end{table}

\subsection{Subgroup Robustness}
\label{sec:subgroups}

\begin{figure*}[t]
\centering
\includegraphics[width=0.85\textwidth]{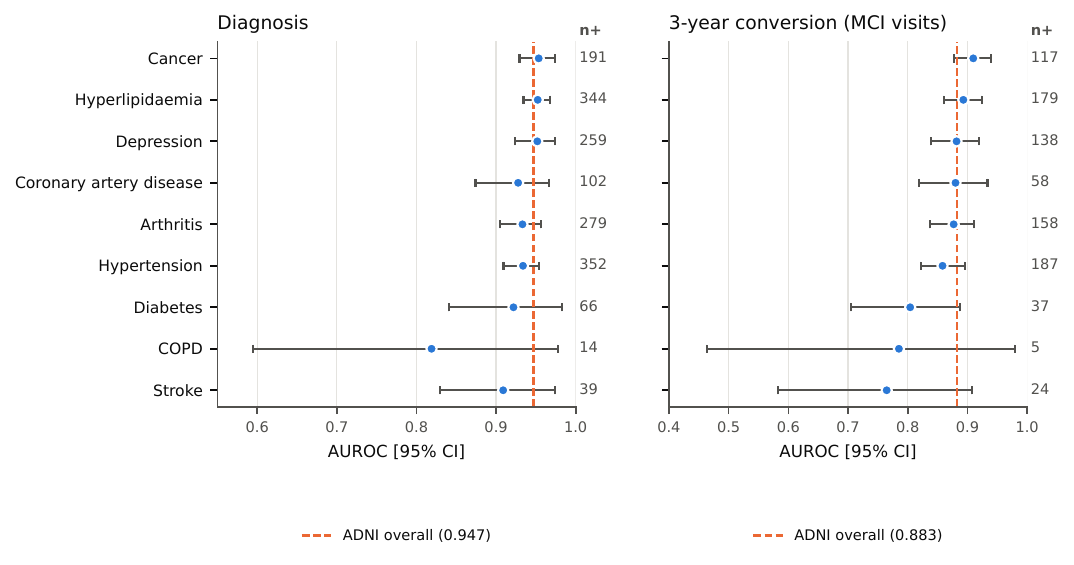}
\caption{Network AUROC within recorded-comorbidity strata, seed 42, for binary diagnosis and MCI-visit conversion. Intervals use 1,000 subject-cluster bootstrap resamples. Numbers at the right of each panel (n+) give the subjects with at least one positive label for that task. Dashed lines show the corresponding overall ADNI AUROC. Strata may overlap, and small subgroups have substantial uncertainty. Point estimates and intervals characterise subgroup performance and its uncertainty.}
\label{fig:subgroup_forest}
\end{figure*}

Table~\ref{tab:subgroups} and Figure~\ref{fig:subgroup_forest} evaluate diagnosis and conversion within each of the nine comorbidity strata (ADNI only, since comorbidity status is unknown in OASIS-2 and MIRIAD). Diagnosis ranges from 0.819 (COPD) to 0.954 (cancer); conversion is more sensitive, with its worst stratum in stroke (0.765; 24 subjects with at least one positive conversion label across 74 positive MCI visits), about 0.12 below the pooled value, identifying stroke as a priority subgroup for further evaluation. The COPD stratum contains only 5 subjects with at least one positive conversion label across 15 positive MCI visits (conversion AUROC 0.785), and its estimates are correspondingly uncertain. Without Group-CVaR, the worst diagnosis stratum is 0.828 (COPD) and the worst conversion stratum 0.768 (stroke), against 0.819 and 0.765 with it; the mean of the three worst strata and the best--worst disparity are also similar.

\begin{table}[t]
\centering
\caption{Subgroup robustness over the nine comorbidity strata (ADNI, seed 42). The stratum is given in parentheses. Seed SD: full model over three seeds.}
\label{tab:subgroups}
\scriptsize
\setlength{\tabcolsep}{2pt}
\resizebox{\columnwidth}{!}{%
\begin{tabular}{lcccc}
\toprule
 & \makecell{Worst-stratum\\diagnosis} & \makecell{Worst-stratum\\conversion} & \makecell{Hardest\\comorbidity} & \makecell{Comorbidity\\mean} \\
\midrule
Full model & 0.819 (COPD) & 0.765 (stroke) & 0.512 (diab.) & 0.565 \\
w/o CVaR & 0.828 (COPD) & 0.768 (stroke) & 0.523 (hyperl.) & 0.565 \\
Seed SD & \(\pm\)0.008 & \(\pm\)0.029 & \(\pm\)0.006 & \(\pm\)0.010 \\
\midrule
\multicolumn{5}{l}{Mean of 3 worst strata: diagnosis 0.883, conversion 0.785 (full); 0.889, 0.799 (no CVaR)} \\
\multicolumn{5}{l}{Disparity (best \(-\) worst): diagnosis 0.135, conversion 0.145 (full); 0.123, 0.144 (no CVaR)} \\
\bottomrule
\end{tabular}}
\end{table}

\subsection{Interpretability}
\label{sec:interpretability}

Network input dependence is assessed with Integrated Gradients (IG) \cite{sundararajan2017axiomatic} and group occlusion. Network explanations use the first batches of each validation loader, capped at 250 subjects (the batch boundary can extend this to 256). This subset comprises ADNI participants without raw MRI. The network explanations therefore describe a sampled ADNI tabular population rather than all three cohorts. MRI embeddings are cached and held fixed during IG, which is computed for diagnosis and forecast input tensors relative to a zero-input baseline. The scalar target is the sum of eligible-output logit contrasts; the attribution describes aggregate model dependence and is not a causal effect or an explanation of one selected visit. Progression explanations include eligible progression-labelled visits and are not restricted to the MCI population used for the primary conversion metric.

\subsubsection{Group Occlusion}
\begin{figure*}[t]
\centering
\includegraphics[width=\textwidth]{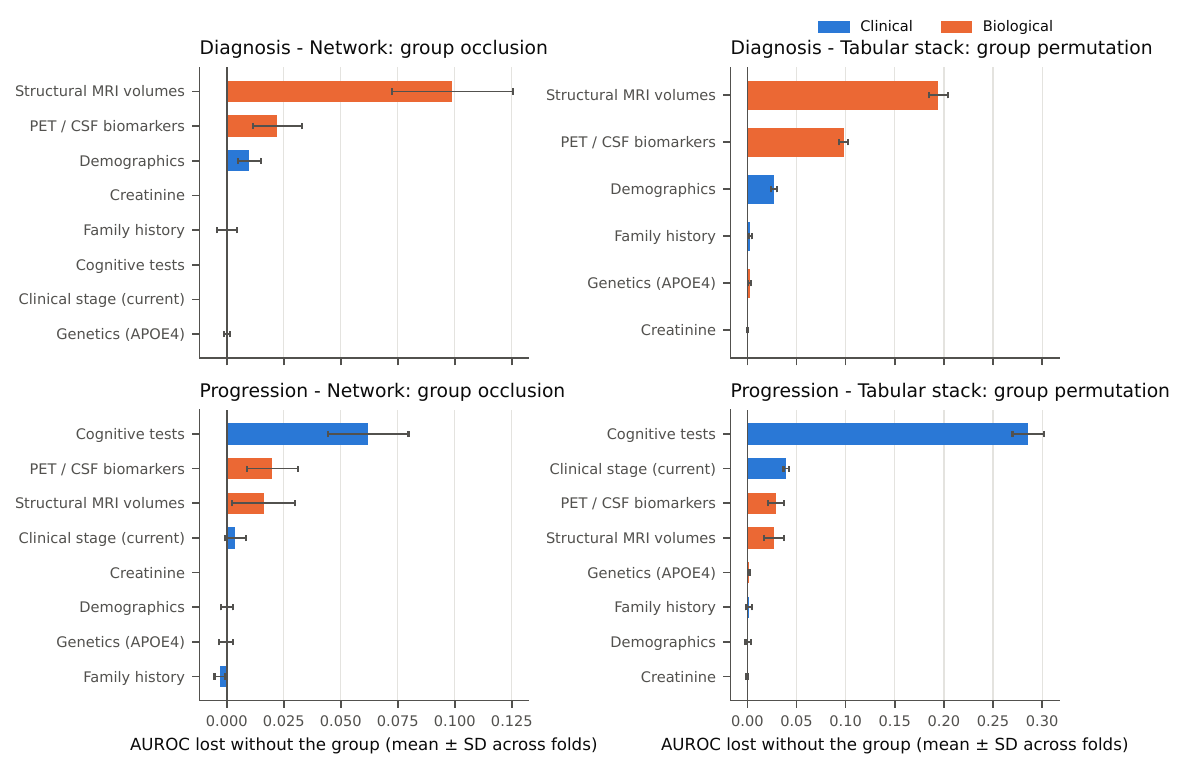}
\caption{Input-group perturbation for diagnosis (top) and eligible-visit progression (bottom): network occlusion (left) and tabular-stack permutation (right), seed 42. Bars and error bars show the fold mean and sample SD. Positive values mean that perturbation lowers AUROC; negative values mean that it increases AUROC. Network explanations use a capped ADNI subset; the tabular panels can use different eligible rows, so magnitudes should not be compared directly. Progression explanations are not MCI-only. Structural MRI volumes are tabular measurements; raw MRI is not evaluated in this subset.}
\label{fig:xai_groups}
\end{figure*}

Occlusion measures how much AUROC is lost when a whole input group is removed (set to ``never measured''), which examines joint dependence on a feature group but can introduce out-of-distribution missingness and remains sensitive to correlated groups (Figure~\ref{fig:xai_groups}). For \textbf{diagnosis}, removing the structural MRI volumes costs 0.099\(\pm\)0.027 AUROC (mean\(\pm\)SD over folds) and removing PET/CSF biomarkers 0.022\(\pm\)0.011. For \textbf{conversion}, removing cognitive tests costs 0.062\(\pm\)0.018, PET/CSF 0.020\(\pm\)0.011 and structural volumes 0.016\(\pm\)0.014. The raw-MRI perturbation is not informative in this subset because these subjects lack raw MRI; it is omitted from the figure. The two methods assess different forms of input dependence. Network occlusion replaces a group by its missing-input representation, whereas tabular permutation disrupts its correspondence with the outcome. A negative bar, such as family history in the network progression panel, means that AUROC increased after perturbation. It does not imply that family history protects against disease. This pattern can reflect correlated inputs, sampling variability or perturbation sensitivity. Fold SD describes variability, not a confidence interval or a significance test.

\subsubsection{Top Features}
\begin{figure*}[t]
\centering
\includegraphics[width=0.92\textwidth]{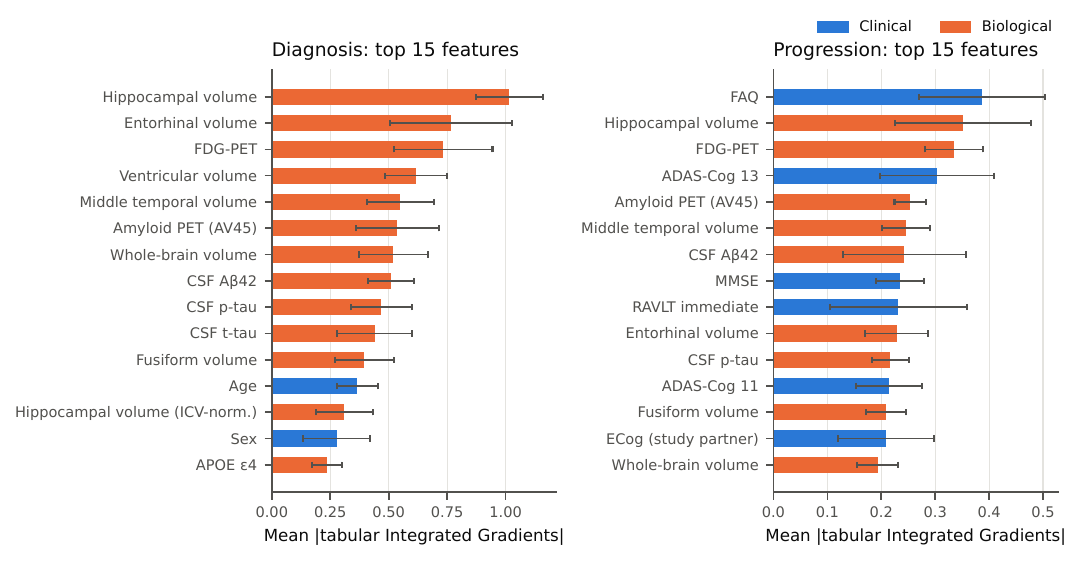}
\caption{Tabular features ranked by mean absolute IG for diagnosis (left) and eligible-visit progression (right); bars and error bars show the equally weighted fold mean and sample SD. The IG target sums eligible-output logit contrasts, with raw MRI held fixed. Related value, availability and history features are grouped by base feature. Absolute IG shows the magnitude of dependence, not whether a feature raises or lowers predicted risk; related inputs can share attribution.}
\label{fig:xai_top}
\end{figure*}

The top-ranked individual features include structural, metabolic and fluid measures associated with AD (Figure~\ref{fig:xai_top}). For \textbf{diagnosis}, the ranking is led by hippocampal and entorhinal volume and FDG-PET, followed by ventricular and middle-temporal volume, amyloid PET (AV45), whole-brain volume, CSF A\(\beta_{42}\), p-tau and t-tau, and fusiform volume. These are the established medial-temporal, metabolic and amyloid/tau markers of AD. For \textbf{conversion}, the ranking is led by the Functional Activities Questionnaire (FAQ), followed by hippocampal volume, FDG-PET, ADAS-13, AV45, middle-temporal volume, CSF A\(\beta_{42}\), MMSE and RAVLT immediate recall, i.e.\ early loss of instrumental activities of daily living and episodic memory, supported by neurodegeneration and amyloid markers. Attribution magnitudes identify the inputs the model relies on, not causal effects.

\section{Discussion}
\label{sec:discussion}
\subsection{Main Findings}
A single leakage-controlled model jointly performs diagnosis, conversion forecasting, cognitive and atrophy prediction, and comorbidity prediction, with diagnosis and stage-1 progression AUROCs of 0.935 and 0.884 across three seeds and low seed variation. The sampled ADNI tabular explanations identify dependence on features associated with AD: diagnosis relies on hippocampal and entorhinal volume, FDG-PET and amyloid and tau markers, whereas conversion relies on functional decline (FAQ) and episodic memory, supported by the same biomarkers.

Part of this division of evidence is imposed by the branch-specific input rules, and the attributions describe model associations rather than causal mechanisms.

The contribution of this work is the implemented multimodal multitask framework and its evaluation: leakage-controlled integration of heterogeneous cohorts, tasks and modalities, comorbidity prediction without condition-defining records, stratified subgroup analysis and external validation. The paired ablations characterise the sensitivity of this configuration to its components; observed differences in pooled AUROC were small (at most 0.004 in absolute value) for the selected endpoints and seed, and the longitudinal contrastive loss was associated with higher comorbidity discrimination.

\subsection{Continuous-Time Modelling}
With elapsed time entering the GRU gates in both temporal models, no statistically detectable AUROC difference was observed between ODE integration and a time-aware GRU in this configuration. The ODE additionally provides a latent state defined at any time between visits, which supports future work on forecasting at chosen horizons.

\subsection{Comorbidity and Modality Contributions}
Auxiliary comorbidity modelling and subgroup evaluation provide complementary views of the clinical setting. The head predicts nine recorded conditions without receiving their defining medical-history, medication or laboratory records, achieving a mean AUROC of 0.571 over three seeds. Recorded conditions also define training strata for Group-CVaR and support condition-specific diagnosis and progression analyses. This design integrates auxiliary supervision and subgroup assessment without using recorded conditions as direct predictive inputs. With and without Group-CVaR, the worst diagnosis and conversion strata were similar (0.819 vs.\ 0.828; 0.765 vs.\ 0.768), and conversion in patients with a history of stroke was the lowest-performing stratum.

The sampled ADNI explanations identify dependence on derived imaging volumes, PET/CSF biomarkers and permitted clinical features. The imaging-focused comparator achieves diagnosis AUROCs of 0.597 on MIRIAD and 0.621 on OASIS-2, compared with 0.424 and 0.629 for the fused network. Together, the cohort and modality analyses characterise how performance varies with input availability and population. The tabular attribution analysis covers ADNI subjects without raw MRI; individual raw-image localisation is outside its scope.

\subsection{Limitations and Future Work}
\label{sec:limitations}

The validation folds also guide model selection, threshold and temperature fitting and hybrid weighting, so internal estimates may be optimistic; external evaluation on OASIS-3 provides an independent assessment. The binary diagnosis and progression endpoints are cohort-defined, and external progression uses the next observed visit. Comorbidity labels are rule-derived from medical histories without independent clinical adjudication. On the internal folds, the tabular stack alone exceeds the network AUROC. Raw MRI is not available for ADNI, and volumes are processed without anatomical registration. Ablations use a single seed. Future work will strengthen the image pathway, harmonise comorbidity labels against clinical records and extend external evaluation to cohorts with matched endpoints.
\section{Conclusion}
\label{sec:conclusion}
This study presents an integrated framework for cohort-defined diagnosis, longitudinal progression, cognitive and volume prediction, and auxiliary comorbidity modelling under missing modalities and irregular observation times. Its contribution combines an adapted SFCN encoder, causal clinical streams, task-aware fusion, ODE-GRU dynamics, task balancing, subgroup objectives and explicit input controls in a common pipeline. Three-seed internal validation achieves mean diagnosis and stage-1 progression AUROCs of 0.935 and 0.884, increasing to 0.951 and 0.909 with the tabular hybrid. Paired ablations characterise the sensitivity of performance to the individual components. Task-specific explanations and comorbidity-stratified evaluation connect the predictive results with their input and population context. External validation on OASIS-3 extends the assessment to another cohort, with hybrid diagnosis and next-visit progression AUROCs of 0.767 and 0.764, improved diagnosis calibration and a next-visit MMSE MAE of 0.86. Together, these experiments provide a structured evaluation of multimodal multitask prediction across heterogeneous Alzheimer's disease research cohorts.

\section*{CRediT authorship contribution statement}
\textbf{Akeem Temitope Otapo:} Conceptualization, Methodology, Investigation, Formal analysis, Data curation and Writing -- original draft.
\textbf{Alice Othmani:} Supervision, Writing -- review \& editing and Project administration.
\textbf{Ghazaleh Khodabandelou (corresponding author):} Conceptualization, Validation and Writing -- review \& editing.
\textbf{Zuheng Ming:} Conceptualization, Validation and Writing -- review \& editing.

\section*{Declaration of competing interest}
The authors declare that they have no known competing financial interests or personal relationships that could have appeared to influence the work reported in this paper.

\section*{Funding}
This work was supported by the Tertiary Education Trust Fund (TETFund), Nigeria [grant number TETF/ES/COE/YABA/TSAS/2022].

\section*{Data availability}
The ADNI, OASIS-2, OASIS-3 and MIRIAD data used in this study are available from their respective repositories to registered users under each cohort's data use agreement. The code is available upon request.

\section*{Acknowledgements}
The authors thank the OASIS, MIRIAD and ADNI investigators for making their data publicly available.


\end{document}